# Natural Language Processing

— A SURVEY —

By
Kevin Mote

May 2002

CptS 499

*"Computer, would you search the Web for references to our company and determine the 3 most common complaints mentioned about us?"*

```
"Sure, Kevin."
```

*"Then summarize and format your findings and insert it into the second page of my slide presentation."*

```
"No problem."
```

*"And also, could you give me a reminder notice a half hour before my plane is supposed to leave?"*

```
"Ok, hold on. . . Yeah, apparently that flight has been delayed.
Looks like you'll have an extra hour. I'll remind you. . . .
Would you like earlier notice if the traffic is bad?"
```

*"Yeah, that'd be great."*

## INTRODUCTION: "Why can't computers understand plain English?"

In the movie *2001: A Space Odyssey* we were introduced to what Arthur C. Clarke (and many computer scientists) believed was an inevitable future companion: a conversant computer. The "HAL 9000 agent" was an artificial agent capable of speaking and understanding English, even reading lips. The Star Trek franchise has kept this dream alive as well. In a classic scene from Star Trek IV, the crew of the Enterprise return in a time-travel mission back to 20$^{th}$ century Earth. In one exchange Scotty asks to borrow a desktop computer while hiding his identity as a time-traveler. He tries in vain to engage the computer by speaking to the screen. The confused owner of the computer taps Scotty on the shoulder and hands him a mouse. Sheepishly, Scotty grabs the mouse, ("Oh, how quaint," he says), thanks the owner, then holds the mouse up to his mouth and says, "Computer, please open the file. . ."

The evidence is becoming increasingly clear that before most of us retire, the presence of conversational computers will be so ubiquitous and their use so universal that we will all smile with amusement at how arcane the "old" modes of input were. The question is, how far off is the arrival of this new age of technology?

Clearly we already have commercial software that can recognize speech, synthesize voices, and even "understand" rudimentary English queries. But how close are we to being able to truly carry on a natural conversation with our desktop computer? ("Computer, would you please determine the most widely recommended health sites on the Web and search them for a discussion of this lower back pain that I describe…")

The pursuit of NLP has a history nearly as long as that of Science Fiction. No sooner had we dreamed of it, then someone started to try to implement it. It has been pointed out that,

> providing the computer with a natural interface, including the ability to understand human speech, has been a research goal for almost 40 years. Speech recognition research started with an attempt to decode isolated words from a small vocabulary. As time progressed, the research community began working on large-vocabulary and continuous-speech tasks.[1]

By any realistic estimate, we have not arrived – but we are getting closer.

> Practical versions of such systems have become moderately usable and commercially successful only in the past few years, however. Even now, these commercial applications either restrict the vocabulary to a few thousand words, in the case of banking or airline reservation systems, or require high-bandwidth, high-feedback situations such as dictation, which requires modifying the user's speech to minimize recognition errors.[2]

In their ground-breaking book on this entire field, Daniel Jurafsky and James Martin explore these issues in detail. They begin their discussion by describing the minimal requirements for creating the language-related parts of a Hal-type computer:

> Minimally, such an agent would have to be capable of interacting with humans via language, which includes understanding humans via **speech recognition** and **natural language understanding** (and, of course, lip-reading), and of communicating with humans via **natural language generation** and **speech synthesis**. HAL would also need to be able to do information retrieval (finding out where needed textual resources reside), **information extraction** (extracting pertinent facts from those textual resources), and **inference** (drawing conclusions based on known facts).[3]

Jurafsky goes on to point out that, although these problems are far from completely solved, much of the language-related technology that HAL needs is currently being developed, with some of it already available commercially. And a great number of scientists at the research labs of Microsoft, IBM, MIT and others, are busily trying to push the technology envelope – and attaining a surprising degree of success. [4]

---

[1] Padmanabhan, Mukund & Michael Picheny, "Large-Vocabulary Speech Recognition Algorithms," *Computer,* Vol. 35 #4 (Apr 2002), p. 42.
[2] Ibid.
[3] Jurafsky, Daniel, and James Martin. *Speech & Language Processing*, p.1.
[4] Ibid., p2.

All of these topics represent a wide range of research interests, but one overarching category that touches each of them is referred to as "*Natural Language Processing"* or NLP. The primary concern of NLP is the effort to find the methods and mechanisms which will allow computers to understand and respond to the natural, colloquial, spoken English language (or any human language).

# 1. THE PURPOSE:

As was already said, the goal of NLP research is to implement within computers the ability to understand (or even converse in) a normal human language. This is to be understood as broader than just an issue of human-computer interfaces. The fictional dialogue at the beginning of this paper illustrates several skills that a conversational computer could make use of. Of course, the user is able to interact with the computer naturally, in common vernacular. But beyond this, notice the more subtle (and significant) skill demonstrated by the computer: It can retrieve *and analyze* pertinent information from a wide variety of sources, many of which are in unstructured formats. In this case it was the Internet and an airline scheduling system, but the world of information exists in uncountable domains: e-mail messages, SQL server databases, archived news reels, PowerPoint presentations, recorded voice conversations, as well as commercial systems of all kinds (banks, hospitals, universities, etc). The future of human-information interaction will demand an unimaginably wide net of data-processing capabilities. But at the central core of all of them will undoubtedly be a *Natural Language Processor.*

The importance of such research is hard to miss. The ever-proliferating amount of information on the web (some estimate its growth rate at nearly 50 terabytes per month!) necessitates more sophisticated methods of retrieval. But in order to access and retrieve *pertinent* information (amidst all the data and garbage) the expedient processor must be able to ferret out the *meaning* of the information desired. And ascertaining *meaning* is one of the central pursuits of NLP research. Another motivation is that as computers become smaller and more ubiquitous, and as people become accustomed to accessing computing resources in many different environments, traditional interface modalities (based on keyboards and monitors) will have to be replaced with modalities that do not depend on such equipment.

Speech interaction is a natural choice for this role. A third motivating factor that could be mentioned is the great numbers of uninitiated computer "*avoiders*". By this I am referring to the segment of society whose access to Web Information is impeded simply by their inability to navigate around keyboards and operating systems. A conversational computer would be a great leap forward for many web-illiterates. Finally (and, perhaps most significantly), the *commercial prospects* for conversational software is staggering. It has already been developed in limited domains (airline booking agents, corporate voice mail, etc), but we clearly have only scratched the proverbial surface in this area.

The study of Natural Language (or "NL") is a burgeoning science and there is an abundance of complex and esoteric literature available. The problem, however, is that it is difficult for the layman, or even the uninitiated scientist, to gain a thorough comprehension of the current state of affairs within this research. Despite the proliferation of research materials (or *because* of it) there remains a barrier to the student who wishes to gain a preliminary understanding of this topic. The goal of this paper, therefore, is describe the purpose, procedures and practical applications of NLP in a clear, balanced, and readable way. We will examine the most recent literature describing the methods and processes of NLP, analyze some of the challenges that researchers are faced with, and briefly survey some of the current and future applications of this science to IT research in general.

Of course, this is a terrifically broad subject addressed in several hundred scholarly articles and numerous large textbooks, so we will need to restrict our scope somewhat. In order to do this we must see how NLP fits into the larger field of Speech, Language and Dialogue Processing.

NLP, defined in its broadest sense, typically encompasses a wide range of modules and methods. Although there is no universal standard to define the scope of an NLP system, in any complete architecture one will usually find the following common elements:

*(1)* ***Speech recognitio****n*: The conversion of an input speech utterance, consisting of a sequence of acoustic-phonetic parame-ters, into a string of words.

*(2)* ***Language understandin****g*: The analysis of this string of words with the aim of producing a meaning representation for the recognized utterance that can be used by the dialogue management component.

*(2)* ***Dialogue Managemen****t*: The control of the interaction between the system and the user, including the coordination of the other components of the system.

*(4)* ***Communication with external system***: For example, with a database sys-tem, expert system, or other computer application.

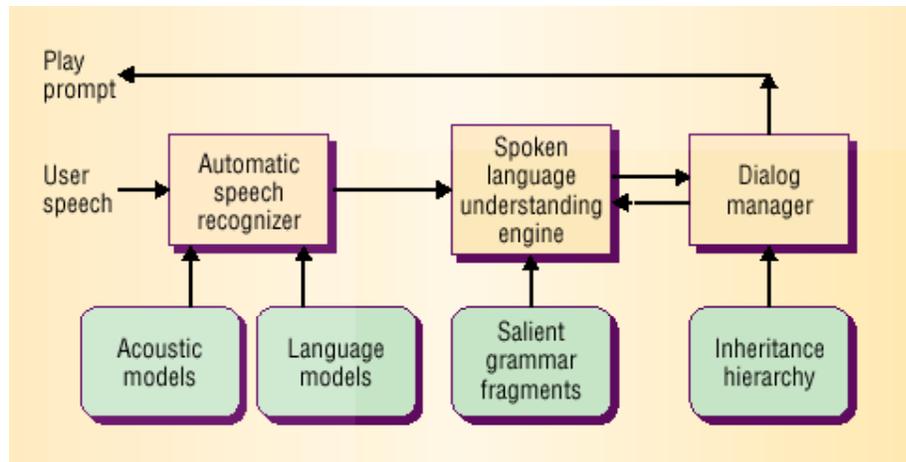

*(5)* ***Response generation***: The specification of the message to be output by the system. *Speech output*: The use of text-to-speech synthesis or prerecorded speech to out-put the system's message.[5]

    Several of these components are illustrated in the diagram on the next page. Each of these topics is fascinating, and could consume an entire research project in its own right. This paper, however, will be focusing primarily on the second component. That is, we will address the specific question: "What are the ingredients and procedures necessary in order to implement a functional *language-understanding* machine? Such a machine will receive as its input a string of tokens from the Speech Recognizer (perhaps in the form of morphemes or "sound units"). The final output at the far end will be the *meaning* of the input statement (described in some standardized notation), ready for a Dialogue manager. In addition to this we may need to look briefly at some of the issues involved in generating a response by the computer (as this skill is required by such NL processes as *verification*), nevertheless, our scope will mainly cover the question of how a computer can understand a human speaker.

    Before we can address the technicalities of this issue, however, it is necessary to first understand the foundation upon which this science is built. Accordingly, the next section will investigate some of the preliminary concepts that give direction and shape to the entire body of research.

---

[5] McTear, Michael. "  "Spoken Dialogue Technology: Enabling the Conversational User Interface,"
    *ACM Computing Surveys.* Vol 34, #1 (Mar 2002): p. 103.

## 2. THE PRELIMINARIES:

In order to appreciate the development that has taken place within the field of NLP, it will be instructive to first examine some of the discussions and debates that have been a sort of "seed bed" from which all the saplings of NL research have grown. In this section we will look at the "psychological" roots of the research, the controversy over methodologies, and the relationship of NLP to other, more fundamental computer science.

### 2.1  Foundation:  "The Human Brain and Cognitive Linguistics"

*"Little can be said with any confidence about 'the architecture of the mind,' and for that reason I (purposely) remain rather obscure on the whole topic."*
*-- Noam Chomsky* [6]

It has been over 50 years since Alan Turing proposed the ultimate test for determining the arrival of truly "intelligent" machines.  The now famous "Turing Test" is set up to evaluate the a computer's *use of language* as the singular criterion for intelligence.  The test consists of a dialogue between an *interrogator,* and two unseen participants. These two participants, one a human and the other a computer, would communicate with the interrogator by teletype.

> The task of the machine is to fool the interrogator into believing it is a person [by its responses]… The task of the other person is to convince the interrogator the other participant is the machine. The critical issue for Turing was that using language as humans do is sufficient as an operational test for intelligence.[7]

Turing predicted that "roughly by the end of the twentieth century a machine with 10 gigabytes of memory would have around a 30% chance of fooling a human interrogator after 5 minutes of questioning."[8] Whether or not this prediction was fulfilled is outside the scope of this paper. But this concept of a computer that could mimic or model conversational humans has continued to influence the minds of many researchers ever since.

---

[6] Stemmer, Brigitte, "An On-Line Interview with Noam Chomsky: On the nature of pragmatics and related issues", in *Brain and Language,* Volume 68, # 3, July 1999, pp. 393-401
[7] Huang, Xuedong, et. al. *Spoken Language Processing: A Guide to Theory, Algorithm, and System Development.* p. 3.
[8] Jurafsky & Martin, p. 7.

And although Turing approached this issue from the point of view of a computer scientist, other branches of science were slowly beginning to pursue the same goal, albeit for a slightly different motivation.

The sciences of Cognitive Theory and Linguistics were interested in devising hypothetical models to describe the inner workings of the brain, and the brain's use of language, respectively.

> In cognitive psychology we are interested in the question, how knowledge is represented internally. What is going on in our minds when we gather and store information about ourselves and about the world around us? How do we retrieve this information, modify it, and use it in everyday life?[9]

> **Technical Terms . . .**
> **"Cognitive linguistics"**
> "Not a single theory but . . . best characterized as a paradigm within linguistics, subsuming a number of distinct theories and research programs. It is characterized by an emphasis on explicating the intimate interrelationship between language and other cognitive faculties. Cognitive linguistics began in the 1970s, and since the mid-1980s has expanded to include research across the full range of subject areas within linguistics: syntax, semantics, phonology, discourse, etc.."[1]
> ___________________
> [1] vanHoek, Karen, in MITECS.

> **Technical Terms . . .**
> **"Psycholinguistics"**
> " Psycholinguistics is the study of people's actions and mental processes as they use language. At its core are speaking and listening, which have been studied in domains as different as language acquisition and language disorders. Yet the primary domain of psycholinguistics is everyday language use."[1]
> ___________________
> [1] Clark, Herbert, in MITECS.

Not surprisingly, these theories and models describing the brain began to grow in sophistication, and soon, as Prof. Hans Uszkoreit explains, "they reached a degree of complexity that [could] only be managed by employing computers."[10] And so, the field that has come to be known as "(theoretical) computational linguistics" was born.

> [Theoretical] computational linguists develop formal models simulating aspects of the human language faculty and implement them as computer programmes. These programmes constitute the basis for the evaluation and further development of the theories.[11]

---

[9] Albert, Dietrich & Josef LukasMahwah, *Knowledge Spaces,* p. 3.
[10] Uszkoreit, Hans. "What is Computational Linguistics?" in an article posted at: http://www.coli.uni-sb.de/~hansu/what_is_cl.html.
11 Ibid. Uszkoreit continues: "In addition to linguistic theories, findings from *cognitive psychology* play a major role in simulating linguistic competence. Within psychology, it is mainly the area of *psycholinguistics* that examines the cognitive processes constituting human language use.

One of the earliest and most influential theoretician in this field was Noam Chomsky. In his widely accepted arguments he maintained that "a significant part of the knowledge in the human mind is not derived by the senses but is fixed in advance, presumably by genetic inheritance."[12] In particular Chomsky argued that this innate structure within the brain facilitated the rapid acquisition of language observed in children despite their so-called "poverty of stimulus" (or the limited linguistic input they receive during their early years). According to this theory, the brain, essentially, is *hard-wired* with a predisposed ability to learn language. This innate structure is referred to as the "Universal Grammar."

One scholar offers this explanation:

> A child's linguistic system is shaped to a significant degree by the utterances to which that child has been exposed. That is why a child speaks the language and dialect of his family and community. Nonetheless, there are aspects of the linguistic system acquired by the child that do not depend on input data in this way. Some cases of this type, it has been argued, reflect the influence of a genetically prespecified body of knowledge about human language. In the literature on generative grammar, the term Universal Grammar -- commonly abbreviated UG -- refers to this body of "hard-wired" knowledge.[13]

It is precisely *this* grammar that computational linguists are now endeavoring to model in software. Drawing on Chomsky's suggestion, these researchers use Finite-State Machines And Context-Free Grammars (described later) to define and describe this linguistic process.

One interesting theory that extends these concepts even further is expressed in a book entitled *NLP and Knowledge Representation*. In it, the authors recognize that natural language is usually considered just "an *interface* or a front end to a system such as an expert system or knowledgebase."[14] In contrast, these researchers embrace the much larger view that natural language is "a knowledge representation and reasoning system whith its own unique, computationally attractive representational and inferential machinery."[15] The most intriguing (and controversial) aspect of this new perspective is its exploration of the

---

…The relevance of computational modeling for psycholinguistic research is reflected in the emergence of a new subdiscipline: computational psycholinguistics."

[12] Manning, Christopher D., and Hinrich Schütze. Foundations of Statistical Natural Language Processing, p. 5.

[13] Pestsky, David, "Linguistic Universals and Universal Grammar," in MITECS, http://cognet.mit.edu/MITECS/Front/linguistics.html.

[14] Iwañska, Łucja M. and Stuart C. Shapiro. Natural Language Processing and Knowledge Representation: Language for Knowledge and Knowledge for Language. Cambridge: The MIT Press: 2000. p. xiv.

[15] Ibid.

relationship between natural language and the human mind. "Taken to an extreme, such approaches speculate that the structure of the human mind is close to natural language. In other words, natural language is essentially the language of human thought."[16]

Unfortunately, as cognitive scientists would all agree, the mysteries and hidden mechanisms of the human brain remain deeply enigmatic. Computational models have made a great deal of headway, especially with the more recent aid of non-invasive neural scanning procedures such as the MRI and PET; nevertheless, the ability to design a computer that can think or act like humans is still a long way off.

At the same time, we can see that the *residual* benefits of CL research have greatly benefited those NL researchers who have been approaching this problem from the field of pure computer science and mathematics. In fact, the convergence of these two approaches of late has produced something of a harmonic amplification of their respective efforts. And the resulting tidal wave of research developments and insights continues to grow.

## 2.2 Methodology: " The Great Debate: Rules or Statistics "

*"But it must be recognized that the notion 'probability of a sentence' is an entirely useless one, under any known interpretation of this term."*
- *Noam Chomsky (1969)*[17]

*"Anytime a linguist leaves the group the recognition rate [of the NL processor] goes up."*
- *Fred Jelinek (then of the IBM speech group) (1988)*[18]

This snappy exchange is characteristic in what is probably the most significant debate amongst the practitioners of NL processing. The question at the center of the debate is simply: "What is the most effective and promising method by which we should attempt to process natural language?" There are two competing camps which answer this question in fundamentally different ways:

---

[16] Ibid.
[17] Chomsky, N. "Quine's Empirical Assumptions", in Davidson, D and Hintikka, J (Eds.), Words and Objections: Essays on the work of WV Quine. P. 57. Quoted in Jurafsky & Martin, p. 191.
[18] In an address to the first *Workshop on the Evaluation of NLP Systems.* Quoted in Jurafsky & Martin, p. 191.

### 2.2.1 The Rationalists: RULES

The first approach, and the one with the longest history, attempts to build a computational NL processor on the foundation of grammatical rules and formal logic. This is seen particularly in the approach of the computational linguists of the previous section. The basic approach actually "has a history that extends back at least 2000 years, but in this [past] century the approach became increasingly formal and rigorous as linguists explore detailed grammars that attempted to describe what were well-formed versus ill-formed utterances of a language."[19]

Accordingly, the algorithms they devise start with simple grammar rules that we all learned in our middle school English classes. Words are connected to build syntactically correct *phrases,* phrases are connected into sentences, and sentences connected to produce dialogues. This entire process is guided by complex and sophisticated sets of rules which define what *is* and is *not* a grammatically correct (or "*well-formed"*) sentence. Hence, this is sometimes referred to as "formal language theory."[20] It is also called "generative linguistics," as these rules are often understood to "generate" a given legal utterance.

### 2.2.2 The Empiricists: STATISTICS

A more recent alternative has begun to compete with the historical one, and it represents a "paradigm shift" in the research field of language processing.[21] (It actually has an early inception, but was largely forgotten due, in part, to the influential arguments of Chomsky.[22]) It has become apparent to many that the basic rules approach is inadequate, as noted by Edward Sapir in his famous quote: "All grammars leak."[23] As Manning and Schütze contend, "It is just not possible to provide an exact and complete characterization of well-formed utterances that cleanly divides them from . . . ill-formed utterances. This is because people are always stretching and bending the 'rules' to meet their communicative needs."[24]

---

[19] Manning & Schütze. *Processing*, p. 3.
[20] Jurafsky & Martin, p. 11.
[21] Juang, "Automatic Recognition", p. 1142.
[22] Jurafsky & Martin, p.14.
[23] Sapir, Edward. *Language: an Introduction to the Study of Speech.* As quoted in Manning & Schütze. p. 4.

The alternative approach offered is one based on *probability theory*. Its seeks to understand language not by grammars but rather by accumulating the statistical characteristics of large *corpora.* It thus tries to determine what the most "likely" interpretation of a given text is, based on a comparison with vast numbers of other similar contexts. The question they ask is "What are the common patterns that occur in language use?"[25]

This paper will not seek to make a judgement on this debate. The ultimate winners of the dispute are by no means obvious yet. In fact, there appears to be a growing collaboration between the disparate groups as both sides continue to recognize the important contributions of the other. In fact, in the most notable recent book on Statistical NLP, the authors concede, "The difference between the approaches is . . . not

> **Technical Terms . . .**
> **"Corpus; Corpora"**
> A "corpus" is large collection of representative examples of a given language in machine-readable text. Such "corpora" (plural form) may consist of samples culled from articles, speeches, plays, or recorded phone conversations. One of the most famous is the 1 million word collection known as the *Brown Corpus*, compiled from newspapers and books in the early 1960's.

absolute but one of degree."[26] Others add that "most methods use the same simple knowledge representation,"[27] and that "methods vary mostly algorithmicly [sic], in ways they derive weights for features in this space."[28]

Although a delineated comparison of the two approaches would definitely be instructive, it is outside the scope of this current project. Instead, this paper will be describing, for the most part, those characteristics of NLP that are common to both approaches (or becoming so). [29]

---

[24] Manning & Schütze, p. 4.
[25] Ibid., p.5.
[26] Ibid.
[27] Roth, Dan, "Learning in Natural Language: Theory and Algorithmic Approaches. p. 1.
[28] Ibid.
[29] There is, in fact, a third approach to NLP that is based on artificial neural networks. It is often referred to as "*Connectionist NLP."* For an introductory article on what was recently the state of the art, see: Christiansen, Morten & Nick Chater, "Connectionist Natural Language Processing: The State of the Art," *Cognitive Science,* Vol. 23 #4 (1999):417-437. Due to the scarcity of research materials regarding this (apparently fledgling) approach, I chose not to address it in this paper.

## 2.3 Structure: " NL Processors as Compilers "

As you read this paper, your goal (one presumes) is to gain an "understanding" of the topic it addresses. For humans, this is a straight-forward job (made more difficult, of course, by "yawn-inducing" subject material). It is, on the other hand, a very daunting task, when considered from the standpoint of a *computational* approach. For the computer, gaining an "*understanding*" (whatever *that* means--more on this later) involves systematically processing through a long chain of related modules, each one providing input for the next. As Paul Gorrell explains,

> One of the reasons that reading a good novel or listening to an interesting lecture can be a pleasurable experience is because we are (blissfully) unaware of the cognitive work we do in understanding individual sentences and relating them to the discourse context. Research in sentence processing investigates the cognitive mechanism (or mechanisms) responsible for the real-time computation of the structural representation that underlies comprehension of visual or auditory language input. Sentence processing involves the rapid integration of various types of information (lexical, structural, discourse, etc.) and research in this area is necessarily interdisciplinary, drawing on work in theoretical linguistics, computer science, and experimental psychology.[30]

To understand this process better, we need to introduce the concept of a "compiler". Anyone who is involved in computer programming understands that when a piece of programming code is written (in most software languages), it cannot be executed in its plain form. Instead, a separate program must first "compile" the code into machine language. "Simply stated, a compiler is a mechanism that reads a program in one language – the source language – and translates it into an equivalent program in another language – the target language."[31] In software development this means changing the code from a "human-readable" form into a "machine-readable" form. This complex and technical process is usually divided (at least conceptually) into several separate "phases." Each phase transforms the source program from one form to another.

---

[30] Gorell, "Sentence Processing", in *MITECS.* http://cognet.mit.edu/MITECS/Entry/gorrell
[31] Aho, *Compilers,* p. 1.

The first phase, known as "**Lexical Analysis**" examines the stream of characters which make up the source program, and groups them into words or "*tokens*." These tokens are then fed into the "**Syntax Analyzer**" which parses the words by grouping them into grammatical *phrases.* At this point, the words still don't "mean" anything to the computer; they are merely strings of tokens, which may or may not be put together in a "legal" form.  The compiler hasn't yet determined whether the program makes any "sense".  It has only evaluated the *form* to find out if it adheres to a pre-determined template. This template is called the "grammar" of the programming language.  The grammar is, essentially, a list of rules that, when properly applied, can create a complete sentence or line of code.

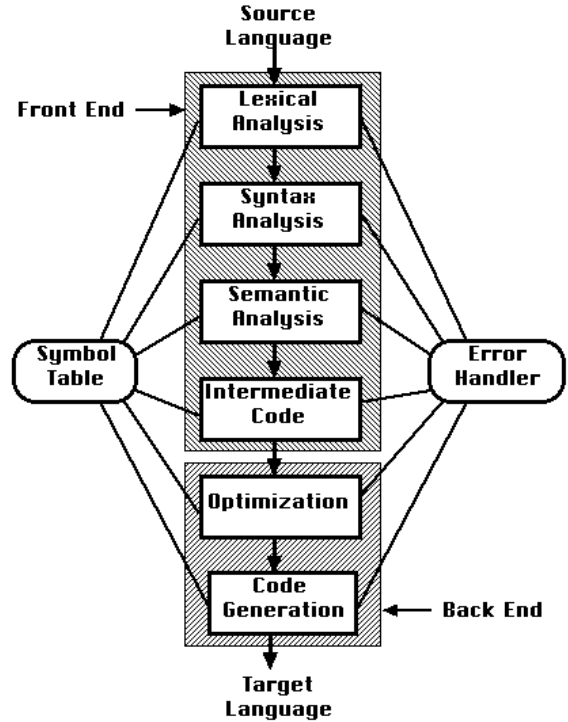

**Phases of a Compiler**

**Source: Nikos Drakos, CS 631 Course Notes ,**
*Programming Language Implementation*

It is the job of the "**Semantic Analyzer**" to actually assign a meaning to the code. It usually accomplishes this by analyzing each syntactically correct phrase, one at a time, in comparison to the grammatical rules which produced it. These rules (or, more properly, "productions") have each been associated with a segment of code – a segment written in *machine language.* The semantic analyzer uses these associations to substitute the human phrases with the machine code.  These three phases form the bulk of the analysis portion of a compiler. [32]

Of course, computer language compilers have been around since the early 1950's and have attained considerable sophistication in their design: systematic techniques for their design and operation have been widely understood and implemented for decades. However, in most of this history these compilers have only been used to translate one *computer code* to another – from a software language into a machine language.

As we shall see, however, these same concepts are directly parallel to the design and conceptual implementation of a natural language processor. In essence, an NL processor is simply a sophisticated compiler – an apparatus that translates a human-readable language into a machine-readable one. And so it is not surprising that an NL processor shares with the compiler the three fundamental phases of translation:

- "Lexical Analysis" – discovering the *words*.
- "Syntax Analysis" – parsing the words according to a *grammar.*
- "Semantic Analysis" – determining the *meaning*.

---

[32] Ibid., p. 10.

Of course, the distinctions between these three areas grow vague in many places. One should not be confused into thinking that they are entirely separate entities. There is a great deal of overlap, both in practical implementation, as well as in the various definitions proposed by NLP researchers. As Robert Dale, in his "Handbook of NLP," explans,

> [Our] attempt at a correlation between a stratificational distinction ([lexicography], syntax, semantics, and pragmatics) and a distinction in terms of granularity (sentence versus discourse) sometimes causes some confusion in thinking about the issues involved in natural language processing; and it is widely recognized that in real terms, it is not so easy to separate the processing of language neatly into boxes corresponding to each of the strata. However, such a separation serves as a useful pedagogical aid, and also constitutes the basis for architectural models that make the task of NL analysis more manageable from a software-engineering point of view.[33]

Therefore, since it is useful to examine each of these modules as intuitively distinct, the remainder of this paper will be dominated by a description of the processes and techniques that each of the phases is comprised of.

---

[33] Dale, Robert. "Symbolic Approaches to Natural Language Processing," *Handbook of Natural Language Processing.* p. 2.

# 3. THE PROCESS:

This section will attempt to "trace a path" through the entire NLP process. It is worth reiterating that this is just one segment of what is usually considered a larger system. But herein we will be assuming that the Automatic Speech Recognizer has already accomplished its job and provided for us some sort of pre-processed input. Likewise, the output at the other end won't be of any particular use without some sort of computational dialogue mechanism which can process it for feedback to the user. But we are interested, here, in the components which constitute the central core of the NLP system.

Accordingly, we will now "lift the hood" on the NLP engine and examine the three fundamental mechanisms which work in tandem to "compute understanding."

## 3.1 Lexical Analysis

*"in the beginning was the word. . ."*
*- John 1:1*

Words are the fundamental building block of language. Every human language, spoken, signed, or written, is composed of words. Every area of speech and language processing, from speech recognition to machine translation to information retrieval on the Web, requires extensive knowledge about words.[34]

> **Technical Terms . . .**
>
> **"Phonemes"**
>
> "The smallest units of speech that serve to distinguish one utterance from another in a language. . . A unit of speech is considered a phoneme if replacing it in a word results in a change of meaning - *pin* becomes *bin*, *bat* becomes *rat*, *cot* becomes *cut*."[1]
>
> ________________
> [1] Lingualinks library Glossary, http://www.sil.org/lingualinks/library/literacy/glossary/cjJ456/krz589.htm

The first step one must take before they can ever hope to "understand" the communication of another is to identify the specific words with which a given statement is composed. In a NL processor, this is the job of the Lexical Analyzer. Lexical Analysis is the process of recognizing a single word from a list of characters, (or, in the case of voice recognition, from an audio spectrograph, or, perhaps, a list of *phonemes)* and to identify certain lexical "features" associated with the word. For a human, of course, this process is trivial. For a computer, however, this takes a fair level of sophistication. The next three subsections will provide an overview of the components that are necessary to accomplish this.

---
[34] Jurafsky & Martin, p. 19.

### 3.1.1  Word Recognition

*"You shall know a word by the company it keeps."*
*- J.R. Firth* [35]

When dealing with *written* communication, the task of identifying individual words is made simple (in most languages) by the common practice of separating the words with "whitespace." This is probably the easiest part of the whole process. That's not to say that there are no challenges, however. As Dale points out, "Not all languages deliver text in the form of words neatly delimited by spaces. Languages such as Chinese, Japanese, and Thai require first that a segmentation process be applied."[36] For the most part, however, the tokenization of written words is a simple routine.

Dealing with *verbal* communication, however, adds another layer of complexity. Here the input is a "continuous speech stream".[37] To identify the lexical units here requires a process known as "*Automatic Speech Recognition"* or ASR.

As has already been mentioned, it is beyond the scope of this paper to describe the ASR process. Needless to say, it is a challenging job, and one with a long history of research. Although current applications have come a long way, the techniques are still far from perfect, and much effort remains to be done. For the rest of this discussion, however, we shall assume that the stream of input text has already been broken up into words and sentences by some ASR mechanism, and the words or "tokens" of the text have been identified.

### 3.1.2  Morphology

---

[35] Quoted in Manning & Schütze, p. 6.
[36] Dale, p. 3.
[37] Ibid.

Once these token-words are ready for subsequent processing, they can serve as input to what is more strictly considered the primary job of the lexical analyzer. As Dale says, "the words are not atomic and are themselves open to further analysis."[38] This is to say that words carry with them lexical *features* by the form in which they appear. In English the word "see" can appear as the word forms "see", "seen", "sees", "seeing", or "saw". Here we enter into the realm of "*morphology.*" A helpful explanation can be found in the MIT Encyclopedia of Cognitive Sciences:

> Morphology is the branch of linguistics that deals with the internal structure of those words that can be broken down further into meaningful parts. Morphology is concerned centrally with how speakers of language understand complex words and how they create new ones. Compare the two English words *marry* and *remarry.* There is no way to break the word *marry* down further into parts whose meanings contribute to the meaning of the whole word, but *remarry* consists of two meaningful parts and therefore lies within the domain of morphology. It is important to stress that we are dealing with meaningful parts. If we look only at sound, then *marry* consists of two syllables and four or five phonemes, but this analysis is purely a matter of phonology and has nothing to do with meaningful structure and hence is outside morphology.[39]

**Technical Terms . . .**

**"Morphology"**

"The study of the way words are built up from smaller…units, [called] *morphemes*. A morpheme is. . . the minimal meaning-bearing unit in a language."[1]

__________________

[1] Jurafsky, p 59.

Hence, the goal in the morphological process is, primarily, to identify *grammatical* features such as parts of speech. Jurafsky offers a helpful distinction between two broad classes of morphemes: ***stems*** and ***affixes***. In most languages, "the stem is the 'main' morpheme of the word, supplying the main meaning, while the affixes add 'additional' meanings of various kinds."[40] Affixes include things like suffixes (as in clear-*ly),* prefixes (as in *un*-clear), and also (in some languages) infixes and circumfixes.

The real job of the morphological process is to determine what the significant morphological units are and to *label* the words accordingly. A common output of such analysis would be a set of data, as in these examples:

[**wolves** (PN) *wolf*]

[**running** (V) (+PRES-PART) *run*]

---

[38] Ibid., p. 4.
[39] Aronoff, Mark, "Morphology", in *MITECS.* http://cognet.mit.edu/MITECS/Entry/aronoff
[40] Jurafsky & Martin, p. 59.

each of which includes: the "*surface*", or "realized" form (e.g., "**wolves**"), the syntactic category ("**PN**" for "plural noun"), and the base word form or stem ("*wolf*").[41] The process of taking a "surface" word and producing the annotated set is known as "*morphological parsing.*" Parsing in general simply means taking some language input and producing some structure for it according to particular rules. So, in order to accomplish this task, we must construct a morphological parser. Such a parser, as typically described, is composed of three separate and essential components: a *lexicon*, a set of *morphological rules,* and a set of *orthographic rules.* We will now look at each of these in turn and illustrate them with one or two examples.

(a) **Lexicon.** This is simply a "repository for words."[42] The simplest conception of a lexicon would be an alphabetic list of *every* single word possible, such as:

…

ask

asked

asker

asking

…

Note in this example that there is a separate entry in this lexicon for every word in every form (i.e., with every possible affix). Of course, such a list would be impractical or even impossible, given the complexity of lexical processes. In Spanish, for example, regular verbs can typically occur in any of approximately thirty five forms; in Finnish one can derive thousands of forms from each lexical item.[43] In practice, therefore, lexical entries typically consist of the stems (and sometimes a separate list of affixes) of a language, together with various other information, depending on the particular theory of grammar.[44]

The choice of whether or not to include affixes comes down to a distinction between two different approaches to lexicography: "Item-Arrangement" (IA) and "Item-Process." The IA model treats complex words as being "composed of two or more atomic morphemes, each of which contributes some semantic, grammatical, and phonological information to the composite whole."[45] And so, to return to a previous example, "wolves" would be composed of two morphemes, the base "wolf", and the affix "-es." Both of these elements would be entries in a lexicon, with the first being listed as belonging to the category N(oun) and the second belonging to a category that takes N's as a base and forms P(lural) N(ouns). In contrast, in the IP model, the lexicon would contain items such as N(ouns) and V(erbs), but elements such as -es "have no existence separated from the morphological rules that introduce them."[46] It is the *rules*, then which govern the formation of new words, and they are the second element in our parser:

    (b) **Combinatorics**. If morphology involves taking words apart to uncover information that can be used in later processing, combinatorics are the rules which describe how they are combined back together. Specifying these rules greatly increases efficiency in terms of storage space, as opposed to the IA practice of "listing every word as an atomic element in a huge inventory."[47] Related to combinatorics is a related study known as "*morphotactics,*" which is a model "explaining which classes of morphemes can follow other classes of morphemes inside a word. For example, the rule that the English plural morpheme follows the noun rather than preceding it."[48]

---

[41] E.g., see Hausser, *Foundations.* p. 243.
[42] Jurafsky & Martin, p. 66.
[43] Sproat, Richard. "Lexical Analysis," in *Handbook of Natural Language Processing.* p. 37. Cf., also, Manning & Schutze, p. 82-83.
[44] Manning & Schultze, p. 266.
[45] Sproat, p. 39.

    Combinatorics often divides morphology into two classes of combination methods: "*Inflectional* morphology," usually involves combining a stem with an affix to produce a related word with, perhaps, a different tense or number, for example. "*Derivational* morphology" combines stems and morphemes to produce a word of a *different* class. The verb *computerize,* for example, can take the derivational suffix *-ation* to produce the noun *computerization*.[49]

    (c) **Orthographic Rules.** The final element in a morphological parser are the set of orthographic or *spelling* rules which are "used to model the changes that occur in a word, usually when two morphemes combine (e.g., the y → *ie* spelling rule. . . that changes city + -s to *cities* rather than *citys*."[50] Although these rules are, in general, fairly straight forward, one must always contend with the preponderance of irregular verbs and exceptions to the rules. For example, we are all familiar with the rule learned in grammar school: "*i* before *e*,. . . except after *c*. . . *except* when they sound like *ay* as in *weight."* But then what do you do with words like *leisure*, or *deity*, or *conscience, or height* (none of which obey the rules *or* the

---

[46] Ibid.
[47] Dale, p.1.
[48] Jurafsky & Martin, p. 65.
[49] Ibid., p. 61.
[50] Ibid.,

exceptions)? Needless to say, the accumulation of rules and special cases can proliferate enormously. And with it grows the complexity of this parsing module as well.

Once we have assembled our morphological parser, we next need a tool or abstract device with which we can *model* the process of lexical parsing. This model will serve as the *design document*, describing how the Lexical Analyzer should work, and illustrating graphically the rules that it should obey. There are many ways to realize this, but the tool most commonly used is known as a "finite-state automaton."

### 3.1.3    Finite Automata

One of the principle tools that is used over and over in many different phases of NLP is a machine known as a "*finite state automaton"* or FSA.  Indeed, Jurafsky calls it "the most important fundamental concept in language processing."[51] And although it is used throughout NLP, its functionality is particularly indispensable in Lexical Analysis.  In this section, we will attempt to describe what a finite state machine *is,* and then, how it is used in the lexical process.

State machines such as FSAs are not actual machines in the sense of being a mechanical object with moving or electrical parts (though it is sometimes helpful to picture it that way).  It is, rather, an abstract idea -- in the same way that the schematic of your VCR that you might find in the user's manual is an abstract design of a working device. But unlike a VCR that reads visual images off a video tape, an FSA reads languages.  Or, more accurately, it *recognizes* them.

Think of it as a "recognizing machine." You might visualize it as a black box filled with cogs and wheels. Whenever the cogs are lined up in any arbitrary configuration, we call that the *state* of the FSA.  Of course there is only so much room in the box, so you can only have a finite number of cogs on your wheels; therefore it is called a *finite* state automaton.

The *input* to the FSA consists of characters of an alphabet. In the case of Lexical Analysis, these characters might be actual ASCII characters used in English words. (In other applications the input alphabet may consist of entire words, or lines from a computer program.) Now, although the words of the English language are composed entirely from this alphabet, that obviously doesn't mean that any random assortment of letters comprises a legitimate word. It is the job of the FSA to determine which words are in the language, and which aren't.

---

[51] Jurafsky & Martin, p. 51.

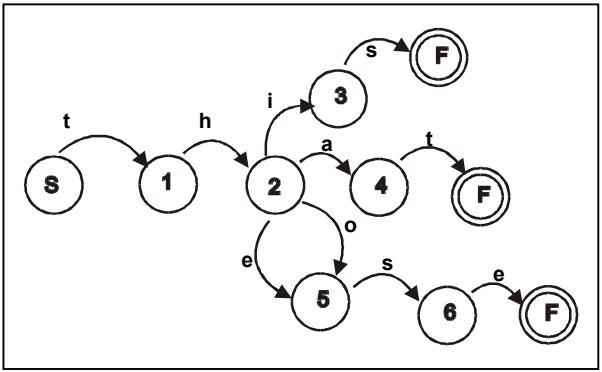

The way this works is that we feed all the letters of a candidate word, one at a time, into the FSA. Each word causes its "cogs" to rotate into a new position or "state." The next input character causes the machine to transition into another new state – one that is entirely dependent on whatever state it was in previously. Once all the characters have been processed, the machine terminates. At that point, the machine may be in a state that has been pre-defined as a legitimate, or "final" state. In that case, the word is "accepted," that is, it is *recognized* as a legal word in the FSAs language. If the machine does not terminate in an "final" state, the word is determined to be illegitimate. (It is not hard to see the usefulness of such a tool in, for example, applications such as spell checking.)

Any machine, of course, must first be designed, and, therefore, every FSA has its corresponding schematic, known as a "state diagram." The notation used in such a diagram consists of numbered **circles** representing the various states that the machine can be in at any particular point, and labeled arrows (or "**edges**") that point from one state to another. These arrows are labeled with the particular character that can cause the automaton to transition from one state to the next. The states, also, are labeled, often just with numbers. In particular, however, the initial state is labeled with an S and the final state with an F.

The diagram at the left is a simple example of an FSA that takes single characters as input. The FSA modeled in this case recognizes the language consisting entirely of the words *this, that, these,* and *those.* Note that from the start state, the only input character that can cause a transition to occur is a *t.* Try to feed it anything else and the FSA will *crash.* If it receives a *t* it transitions to state 1. The next character must be an *h,* upon which it will transition to state 2. At this point a variety of letters may be accepted by the automaton. Eventually, as long as the input word is in the language, all the letters will be processed and the FSA will terminate successfully.

Note also that "the sequence of state changes through which such a machine passes is complete determined by the input stream. This being so, it is clear that, for a given input string, there is a unique path through the machine."[52] For a given string, the FSA will always go through the same set of transitions. For this reason it is called "*deterministic.*"

Now, of course, it is untenable to have a separate FA for every word form in the English language. For this reason we must modify our design by gathering morphemes together, allowing transitions to occur in accordance with morphological or combinatoric rules. Notice, for example, the figure on the right.[53] This FSA assumes that the lexicon includes regular nouns (which take the regular –*s* plural), such as *dog, cat,* and *aardvark.* These, of course, are the vast majority of English nouns. The FSA also accepts irregular noun forms that don't take –*s*, both singular (*goose, mouse*) and plural (*geese, mice*).[54]

It is perhaps not surprising that such diagrams can become extraordinarily complex and involved. They are, nevertheless, a significant and indispensable tool in the belt of lexicologists. There is also, in fact, an important extension of finite-state automata known as finite-state *transducers.* In addition to recognizing words in a language, transducers can also generate output symbols. This is particularly helpful, for example, in modeling orthographic rules. Moreover, another extension to the FSA is 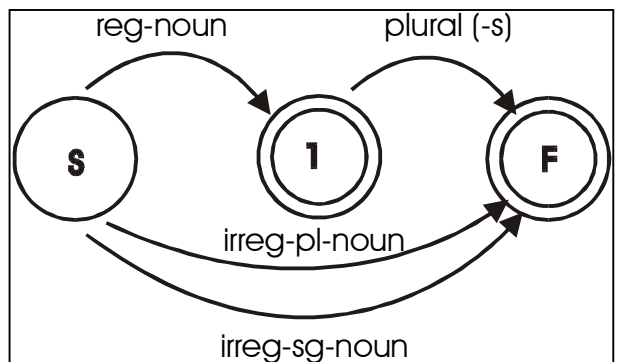 one called a *weighted-* or *probabilistic FSA.* These devices modify the basic FSA by labeling the arrows with the *probability* that a given input word will take that particular path. We will return to these when we discuss Hidden Markov Models (Section 3.2).

---

[52] Moisl, Hermann, "ELL230:Computational Linguistics 1," *Lecture Notes.* Lecture #8, p. 1.
[53] Adapted from Jurafsky & Martin, p. 67.
[54] Ibid. p. 66.

### 3.1.4  Word Prediction: n-grams

Before we leave lexical analysis, one more concept needs to be described because of its nearly universal utility in NLP applications.  It is an example of one of the most important contributions of the *statistical* approach to Lexical analysis. It is used in particular to enable NL processors to *predict* words. Imagine, for example, how helpful it would be to have a device that could actually predict what the next unspoken word in a given conversation will be. At first, such an idea might be dismissed as completely implausible, but, in fact, it is not really all that difficult. By way of example, Jurafsky asks, What word is likely to follow this sentence fragment?

"I'd like to make a collect. . . "

"Probably most of you concluded that a very likely word is *call,* although it's possible the next word could be *telephone,* or *person-to-person. . . .* The moral here is that guessing words is not as amazing as it seems, at least if we don't require perfect accuracy."[55]  Indeed, this is the central principle in probabilistic methods: determining what words are *most likely* to appear in a given context. A related problem is computing the probability of a *sequence* of words. Dr. Allen Gorin and his colleagues at AT&T Labs describe it this way: "The state-of-the-art approach to recognizing unconstrained spoken language involves training a stochastic language model that predicts word sequence probability."[56]

One of the primary methods to actually do such prediction, makes use of concepts known as *n-gram models* or *Markov chains*.  Jurafsky and Martin have an extended explanation that is particularly helpful:

> The simplest possible model of word sequences would simply let any word of the language follow any other word. . . . Each word would then have an equal probability of following every other word. If English had 100,000 words, the probability of any word following any other word would be .00001."[57]

In a slightly more complex model, the *frequency* of a given word would be used to compute its probability.

---

[55] Jurafsky & Martin, p.191.
[56] Gorin, Allen, et. al. "Automated Natural Spoken Dialog," *IEEE Computing*. Vol. 4 #2 (Apr. 2002): p. 52.
[57] Jurafsky & Martin, p. 196.

For example, the word *the* has a high relative frequency, [occurring] 69,971 times in the Brown corpus[58] of 1,000,000 words (i.e., 7% of the words in this particular corpus are *the*). By contrast, the word *rabbit*, occurs only 11 times in the Brown corpus.

We can use these relative frequencies to assign a probability distribution across following words. So if we've just seen the string *Anyhow,* we can use the probability .07 for *the* and .00001 for *rabbit* to guess the next word. But suppose we've just seen the following string:

Just then, the white

In this context, *rabbit* seems like a more reasonable word to follow *white* than *the* does. This suggests that instead of just looking at the individual relative frequencies of words, we should look at the conditional probability of a word given the previous words. That is, the probability of seeing *rabbit* given that we just saw *white* (which we will represent as $P(rabbit|white)$) is higher than the probability of *rabbit* otherwise.[59]

The authors go on to explain how we can use this intuition to compute the probability of a complete string of words. What is the probability of word *w* appearing, given all of the words that we have seen so far? To be more precise, "given a sentence S= $v_1\ v_2\ \ldots\ v_n$, the goal is to estimate the probability of the word $v_i$ given the history of all preceding words: $P(v_l\ |\ v_1\ v_2\ldots v_{l-1})$."[60] Unfortunately, though such computation would have a high likelihood of accuracy, "we don't have any easy way to compute the probability of a word given a long sequence of preceding words. . . . We can't just count the number of times every word occurs following every long string;"[61] we would quickly run out of memory! Or, as Gorin puts it, "Data sparseness makes estimating these probabilities for all possible histories intractable."[62]

Instead, as Jurafsky explains;

We solve this problem by making a useful simplification: we *approximate* the probability of a word given all the previous words. The approximation we will use is very simple: the probability of the word given the single previous word! The **bigram** model approximates the probability of the preceding words $P(w_n|w_{1 \to n-1})$ by the conditional probability of the preceding word $P(w_n|w_{n-1})$. In other words, instead of computing the probability

$P(rabbit|Just\ the\ other\ day\ I\ saw\ a)$

We approximate it with the probability

---

[58] Cf. the definition of "corpus" above.
[59] Jurafsky & Martin, pp. 196-197.
[60] Gorin, p. 52.
[61] Jurafsky & Martin, p.197.

$$P(\text{rabbit}|\text{a})$$

It was Andrei Markov who introduced this simple assumption that we can estimate a word's probability based only on the previous word, or, more generally, that we can "predict the probability of some future unit without looking too far in the past."[63] We can extend this notion of a bigram (which looks at one word in the history) to a trigram (which looks 2 words back), to an *n-gram* (looking *n* words in the past). Such collection of words is sometimes referred to as a *Markov chain.*

Of course, in practice *n* never gets very large. Manning and Schütze point out why:

> If we conservatively assume that a speaker is staying within a vocabulary of 20,000 words, then we get the [following] estimates for numbers of parameters [which we must process]:
>
> 1st order (bigram model):     20,000 x 19,999 = 400 million
> 2nd order (trigram model):    $20,000^2$ x 19,999 = 8 trillion
> 3rd order (four-gram model):  $20,000^3$ x 19,999 = 1.6 x $10^{17}$
>
> So we quickly see that producing a five-gram model. . . may well not be practical. . . .For this reason, *n*-gram systems currently usually use bigrams or trigrams (an often make do with a smaller vocabulary).[64]

In order to actually use such an *n*-gram model, however, it must first be "trained" on some corpus. (In order to compute the probability of any particular bigram appearing in a sentence, we must know how often that bigram appears in general; "training" is the process of counting all the bigrams in a corpus and computing all their probabilities.) Again, this will require an approximation (for we obviously can't count all the times that a given bigram has appeared in every conversation or document in history). Unfortunately, any approximation introduces a flaw into our system because no matter how large the corpus is that we use as our "test-case," there is always bound to be some perfectly legitimate bigram that won't be present. The training corpus is therefore termed *"sparse"* – and the sparseness of a corpus grows as *n* increases.[65]

The problem with this is that there can be any number of legal bigrams (or trigrams) which, because they did not happen to appear in the training corpus, will be given a "zero probability" when they should really have a non-zero probability. In order to resolve this difficulty, the n-gram model is usually modified (after it has been trained) by a process known as "smoothing." Smoothing is the "task of reevaluating some of the zero-probability and low-probability *N*-grams, and assigning them non-zero values."[66] Quite a number of techniques have been developed to accomplish this. The list includes the simplistic "add-one" method (which essentially gives every possible bigram at least a one-in-*n* probability), the "Witten-Bell Discounting" (which, in concept, uses "the count of things you've seen *once* to help estimate the count of things you've *never* seen"[67]), the "Good-Turing Discounting" method (which computes the frequency of *n*-gram frequencies), as well as various "Back-off" methods (which start with the trigram model, for instance, and then "back-off" to a bigram model when the trigram has a zero-probability.) And, of course, active research continues to search for better ways to deal with sparse data.[68]

---

[62] Gofin, p. 52.
[63] Jurafsky & Martin, p. 197. Note: this so-called "Markov assumption" is also used in *probabilistic automatons*, in which we can safely predict that the next state in the automaton is dependent on a finite history.
[64] Manning & Schütze, p. 193-194.

The advantages of the n-gram model are, perhaps, obvious. "In speech recognition and understanding, a probabilistic *N*-gram FSG has found widespread use due to its implementational ease and its consistency with the structure of [other popular statistical methods]."[69] And combined with such tools as Bayesian Inference (see sidebar on next page) it can be an extremely powerful device.

Through the techniques discussed above (and, of course, many others) we are now able to translate a string of text-characters or a stream of acoustic-phonemes into words (and their associated features), a process also known as "*chunking*". We have seen that the three primary elements that are necessary to accomplish this are: (1) a **lexicon** listing all the stems in the language, as well as affixes, and other lexical data, (2) **morphological rules** or *morphotactics* which explain how the word "chunks" or morphemes may be glued together, and (3) **orthographic** or **spelling rules** which describe how the spelling of certain stems change as they are added to their affixes.

The form of the output from this process varies, depending on the implementation, but typically the output of the Lexical Analyzer will be a string of words, each one associated with certain lexical information. This output from one module in our Natural Language Processor now becomes input for the next: The Syntax Analyzer.

---

[65] Cf. Gorin, p. 52.
[66] Jurafsky & Martin, p. 207.
[67] Jurafsky & Martin, p. 211.
[68] Manning & Schütze, p. 224.
[69] Juang, Biing-Hwang & Sadaoki Furui. "Automatic Recognition and Understanding of Spoken Language—A First Step Toward Natural Human-Machine Communication," *Proceedings of the IEEE*, Vol. 88, #8 (Aug 2000): p. 1156.

## A Clever Idea . . .

### "Bayesian Inference"

Not all NLP ideas have been born in the computer age. One particularly useful technique was developed in the 1700's by an English Reverend named Thomas Bayes. Despite its "ancient history" Bayesian classification is used throughout NLP in such jobs as speech recognition, spell-checking, OCR, part-of-speech tagging and probabilistic parsing. One lucrative NL software company in England bases its entire product line on Bayesian statistics.[70]

Bayes' technique is to decide, given an observation, which of a set of classes it belongs to. For speech recognition, the observation may be an aural spectrograph or a string of phonemes making up the sound of a word. Our job is to classify which 'lexeme' the string corresponds to. First we must consider all possible words.[71] Out of this universe of words we want to choose the word which is most probable, given the observation. In statistical notation this probability is written:

$$P(word|observation)$$

We want to find the single word which has the highest probability of being the *right* word. But how is this probability computed? The clever trick is to use a statistical rule, developed by Bayes, which allows us to replace a hard computation P(w|O) with three easier ones according to this equivalency:

$$P(w|O) = \frac{P(O|w)P(w)}{P(O)}$$

The probabilities on the right-hand side are easier to compute than our original problem on the left-hand side. For example, P(w), the probability of the word itself can be estimated by the frequency of the word. In fact, P(O) is the only difficult one to estimate, but because we are comparing all possible choices for a given observation, P(O) is always the same and therefore, it drops out. So our problem reduces to a simple equation: Our best guess for the right word, $\hat{w}$, is the word which has the highest product of two probabilities. Thus,

$$\hat{w} = \max P(O|w) P(w)$$

---

[70] The company is "Autonomy." Michael Lynch, it's CEO, is interviewed by Jerry Borrell in "Language Comes Alive", *Upside* (May 2001), pp. 38-44.
[71] This explanation is based on the treatment in Jurafsky, pp147-149.

We simply compute this equation for each word and choose the one which has the largest product. The beauty of this technique is that it improves its performance with experience. The more words it processes, the more accurate its computations become. In essence, a Bayesian computer can actually "learn by reading."[3][72]

## 3.2  Syntax Analysis

Syntax Analysis is considered to have the most well-established techniques in the field of NLP.[73] Moreover, current research in this field continues to be in a state of dynamic development.

> Probably there are very few fields in cognitive science that have shown as distinct a growth in the last decade as the field of syntax acquisition. The increase in preciseness and knowledge has been extremely large, so large as to make the field hardly recognizable when compared to work of much more than a decade ago.[74]

In this section we will dissect this module of NLP by describing the primary goals, features and methods used in the syntactic process.  We begin by defining what it *is.*

Simply put, Syntax Analysis is focused on the study of *sentences.*  If Lexical Analysis involved the process of breaking a composite word down into its constituent parts, Syntax Analysis involves the process of breaking a *sentence* down into its constituent parts.  Lexical Analysis was concerned morphological and orthographic rules. Now,  we are primarily interested in the *grammatical* rules that hold a sentence together, including, for instance, "parts-of-speech" (or POS). As Manning & Schütze explain,

> Words do not occur in just any old order. Languages have constraints on *word order*. . . [Furthermore,] words in a sentence are not just strung together as a sequence of parts of speech, like beads on a necklace. Instead, words are organized into *phrases*, groupings of words that are clumped as a unit. *Syntax* is the study of the regularities and constraints of word order and phrase structure.[75]

---

[72] Lynch, quoted in Borrell, p.39.
[73] According to Dale, p. 4.
[74] Wexler, Kenneth. "Acquisition of Syntax," in MITECS, http://cognet.mit.edu/MITECS/Front/linguistics.html

Parts-of-speech, such as nouns, verbs and adjectives, have been introduced earlier. We saw in the previous section that morphological parsing requires some knowledge of POS in order, for example, to make use of the lexical rule that most *verbs* can receive the gerund suffix *–ing*, whereas most *nouns* can receive the plural suffix *–s*.

With Syntax Analysis, however, the determination of POS is fundamental and central to the entire process. We will use these POS for "parsing" a sentence, by which we mean, "the process of analyzing a sentence to determine its syntactic structure according to a formal grammar"[76]. As Williams puts it, "Syntax is the study of the part of the human linguistic system that determines how sentences are put together out of words. Syntax interfaces with the semantic and phonological components, which interpret the representations ("sentences") provided by syntax."[77] Interestingly, he also points out that "it has emerged in the last several decades that the system is largely universal across languages and language types, and therefore presumably innate, with narrow channels within which languages may differ from one another (Chomsky 1965; 1981b), a conclusion confirmed by studies of child language acquisition (Crain 1990 and numerous others)."[78]

To understand this process, it will be helpful, again, to break it down into several elements or sub-activities. These elements and concepts are all necessary components in Syntax Analysis, both the foundation on which it is built, and the methods which it employs. The elements that will be explained in this section include (1) *Part-of-Speech Tagging*, which labels the words with their syntactic categories, (2) *Defining Structure*, which uses "Context-Free Grammars" to model syntactic word combinations, and (3) *Parsing,* which refers to the algorithmic processes which analyze the structure of a given string of words.

---

[75] Manning & Schütze, p. 93.
[76] Samuelsson, Christer, & Mats Wirén. "Parsing Techniques," in *Handbook of Natural Language Processing,* Robert Dale, et. al. (eds.), p. 59.
[77] Williams, Edwin. "Syntax," in MITECS, http://cognet.mit.edu/MITECS/Front/linguistics.html.
[78] Ibid.

### 3.2.1 POS Tagging

> **Technical Terms . . .**
> **"Tag"**
> An abbreviated label corresponding to a part-of-speech.
> **"Tagsets"**
> A list of all tags used in a particular tagging system.

The goal of research on NLP is to parse and understand human language. Unfortunately, we are still far from achieving this goal. For this reason, "much research in NLP has focussed on intermediate tasks that make sense of some of the structure inherent in language without requiring complete understanding. One such task is part-of-speech tagging, or simply *tagging*."[79]

Tagging is simply the process of appending every word in a sentence with a "tag" representing its part of speech. The/DET process/NN is/AUX quite/ADV simple/ADJ, as/CONJ this/DET sentence/NN illustrates/VB. In order for an NL processor to make use of corpora (such as the Brown corpus mentioned earlier), the corpora must first be tagged in their entirety.

Before we can accomplish this process we must first establish a *standard* set of tags (a *tagset*) which all processors can refer to in assigning or deciphering a given corpus' tags.

One of the most influential and widely used tagsets is the 87-tag tagset used for the Brown corpus.[80] An abridged version of the Penn Treebank set, which is based on Brown, is shown in the table on the right[81]. Although not

| Tag | Description | Example |
|---|---|---|
| CC | Coordin Conjunction | *and, or* |
| DT | Determiner | *a, the* |
| IN | Preposition | *of, in, by* |
| JJ | Adjective | *red, big* |
| MD | Modal | *can, should* |

seen in the portion listed here, it even includes punctuation.

Once we have a tagset and a tag-trained NL processor, we can begin the first step in parsing a sentence: to analyze each word (in context, of course – more on this briefly), and to determine a single best tag for each word. In some cases this is a straightforward task, as most words in the English language can only belong to one syntactic category. We quickly discover a problem, however, when we encounter an *ambiguous* statement. Consider the commonly cited example:

---

[79] Manning & Schütze, p. 341.
[80] Jurafsky & Martin, p. 296. Other popular sets include the 45-tag Penn Treebank set, and the 61-tag CLAWS5 tagset, also known as "C5" developed at the Univ. of Lancaster.
[81] Adapted from Jurafsky & Martin, p. 297.

>They are flying planes.

This simple sentence is ambiguous on more than one level, but notice particularly the word "flying." Should it receive the /VB tag as a verb, or the /JJ tag as an adjective modifying the noun "planes"? *Resolving* such ambiguities is one of the central problems of POS tagging.

In the face of this problem, two predominant methods have been developed (corresponding to the two primary methods of NLP[82]): *rule-based* tagging and *stochastic* or *probabilistic* tagging.

Jurafsky offers this distinction:

> Rule-based taggers generally involve a large database of hand-written disambiguation rule [sic] which specify, for example, that an ambiguous word is a noun rather than a verb if it follows a determiner. . . . Stochastic taggers generally resolve tagging ambiguities by using a training corpus to compute the probability of a given word having a given tag in a given context.[83]

Rule-based taggers – also called "deterministic" taggers, because the rules *determine* which tag a word should receive – have a history stretching back to the early 1960's. They have proven, however to be extraordinarily challenging. Manning & Schütze describe just one of the ambiguities that make it so difficult:

> Many content words in English can have various parts of speech. For example, there is a very productive process in English which allows almost any noun to be turned into a verb, for example, "*Next, you **flour** the pan*", or, "J*ust **pencil** it in".* This means that almost any noun should also be listed in a dictionary as a verb as well, and we lose a lot of constraining information needed for tagging.[84]

An alternative to this approach (and one with an equally long history), is to apply the science of probabilities to the process of tagging, and to observe, for example, that "although *flour* can be used as a verb, an occurrence of *flour* is much more likely to be a noun."[85] To put it another way,

> The distribution of a word's usages across different parts of speech is typically extremely uneven. Even for words with a number of parts of speech, they usually occur used as one particular part of speech. . . . Indeed, this uneven distribution is one reason why one might expect statistical approaches to tagging to be better than deterministic approaches: in a deterministic approach one can only say that a word can or cannot be a verb, and there is a temptation to leave out the verb possibility if it is very rare (since doing so will probably lift the level of overall performance), whereas within a statistical approach, we can say that a word has an extremely high *a priori* probability of being a noun, but there is a small chance that it might be being used as a verb or even some other part of speech.[86]

---
[82] See Section II-2 "The Great Debate", above.
[83] Jurafsky & Martin, p. 300.
[84] Manning & Schütze, p. 343.
[85] Ibid.

That being said, it must be pointed out that there is no consensus on the best approach to this problem. Disadvantages of the statistical methods can be observed as well. Hausser notes, for example, that "if it turns out that a certain form has been classified incorrectly by a statistical tagger, there is no way to correct this particular error." In contrast, he explains, a rules-based system is a more "solid" solution, because "if a word form is not analyzed . . . correctly in such a system, then the cause is either a missing entry in the lexicon or an error in the rules. The origin of the mistake can be identified and corrected, thus solidly improving the recognition rate of the system."[87]

Another advantage of the rules-based approach has to do with *hapax legomena* (words that only appear once in a corpus). Hausser notes,

> For statistics, the hapax legomena constitute the quasi-unanalyzable residue of a corpus. A rule-based approach, in contrast, can not only analyze hapax legomena precisely, but can reduce their share at the level of elementary base forms by half [by breaking compound words which only appear once into their constituent parts, which appear more often].[88]

Despite the uncertainty of the experts regarding the preferred method of tagging (or rather, *because* of it), the tagging process continues to develop as a combination of the two approaches. Indeed, here is another one of the many areas in which the convergence of disparate approaches can be observed.

Tagging, of course, does not give us the whole story. It is rather "an intermediate layer of representation that is useful and more tractable than full parsing."[89] And "even though it is limited, the information we get from tagging is still quite useful. Tagging can be used in information extraction, question answering, and shallow parsing."[90]

---

[86] Ibid., p. 344.
[87] Hausser, p. 299.
[88] Ibid.
[89] Manning & Schütze, p. 343.

### 3.2.2  Defining Structure – Context Free Grammars

In the English language, words can often be grouped together into phrases. These phrases (known as "*constituents"*) are sub-units of a sentence.  A "noun phrase", for example, is a group of words containing at least one noun that are grouped as a unit. Such units can sometimes be identified by their transiency in a sentence. For example,

"***This noun phrase*** can be found in various locations of this sentence."
"You can find ***this noun phrase*** in various locations of this sentence."
"You can find, in various locations of this sentence, ***this noun phrase***."

Other phrase types include verb phrases ("*can be found*"), prepositional phrases ("*of this sentence*")*,* as well as adjective or adverb phrases.

Most of us can still remember our middle-school English grammar class in which we were taught to dissect and identify these phrases. We did so (or tried to at least) by learning the various rules in our Grammar textbook. The goal of Syntax Analysis is, essentially, to teach these same rules to the computer.  This book of rules, in NLP, is known as the "*Context-Free Grammar"* or *CFG  (*This is a rather obscure and unhelpful name, having its roots in historical computer theory, so it is often referred to instead as a "*phrase structure grammar*" because it describes the structure of phrases.)

Using these rules we can examine a sentence to determine whether or not it is "grammatically correct."  A common method used to perform this analysis (in those English classes at least) was an exercise known as "diagramming the sentence." In a similar manner, computational linguistics analyze sentences using what is known as a "*parse tree*", such as the one shown at the left. What this tree illustrates is that a "legal" sentence can be made of a noun phrase and a verb phrase. Each of these phrases can be further sub-divided into additional "branches" of the tree. The actual words of the sentence are located at the very end of the branches (called the "*leaves*" of the tree). The joints in the tree (called "nodes") are labeled with abbreviations for the constituent types or the parts of speech. The top of the tree is called the "root" (so we have a tree that is growing "upside-down" ).

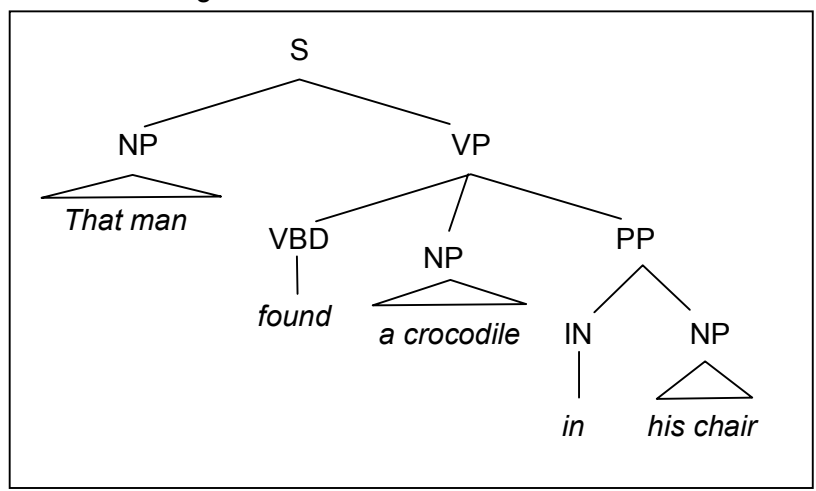

---

[90] Ibid.

In order to encapsulate these rules in an efficient way, a particular notation is usually adopted which lines up all the rules into a list. All the rules in the list take the form:

S → NP VP

which reads, "a S(entence) can take the form of a N(oun) P(hrase) followed by a V(erb) P(hrase). The rules are called "*productions.*" For another example, consider this one:

| 1.1.1.1  Productions | 1.1  Examples |
|---|---|
| S → NP VP | I + want a morning flight |
| NP → *Pronoun*  <br> \| Proper-Noun  <br> \| Det Nominal | I  <br> Los Angeles  <br> a + flight |
| Nominal → Noun Nominal  <br> \| Noun | Morning + flight  <br> Flights |
| VP → Verb  <br> \| Verb NP  <br> \| Verb NP PP  <br> \| Verb PP | Do  <br> Want + a flight  <br> Leave + Boston + at night  <br> Leaving + on Thursday |
| PP → Preposition NP | From + Los Angeles |

Det → a

Det → the

In this case, the right-hand side of the production contains actual words from the language (known as "*terminals,*" because the tree branches *terminate* when they get there). The symbol on the left-hand side (called a "*non-terminal*") can, in this case, produce two possible terminals. (The short-hand notation for this is a vertical bar "|" which is read "or"). Needless to say, a list of all the productions needed to describe a language can grow exceedingly long. A very abbreviated grammar can be seen in the accompanying figure.[91]

Of course, the task of creating these lists isn't technically a part of Syntax Analysis proper. Rather, it must be performed in advance "*by hand*" – which is to say that linguists must accomplish the arduous task of formulating all the rules necessary to describe all grammatically correct sentences.  Obviously such a task is never complete (and, some would argue, never completable) and the number of rules in practical applications often exceeds several thousand[92]. Nevertheless, context free grammars are considered to be extremely powerful in their ability to *recognize* most sentences in a human language -- in virtually all of the sophisticated and complex forms that those sentences can take.[93] Of course we will always have the problems of "*undergeneration," (*where one's grammar is unable to provide an analysis for a particular sentence), as well as "*ambiguity*" (where multiple analyses are provided),[94] but in most cases these problems can be rectified by making improvements to the grammar. The underlying assumption in all this is

> That natural languages as a class are context free, that is, capable of being generated by context free grammars. This is not strictly true – there are a very few isolated instances of NL constructions which cannot be so generated – but it is certainly true of the proverbial 99.999...% of NL syntactic structures, and that's good enough for us.[95]

---

[91] Adapted from Jurafsky & Martin, p. 330.

Once a large enough subset of the language has been modeled by a CFG, we are ready to enlist the aid of computers to perform the actual process of analyzing sentences -- an activity known as "*parsing.*"

### 3.2.3    Parsing

In order to make use of the Grammar so meticulously constructed by the linguists, an NL processor must be able to take the string of words handed to it by the Lexical analyzer, and *match* those words to the corresponding productions in the CFG.  "A parser is a recognizer that produces associated structural analyses ([e.g.] parse trees) according to the grammar."[96] It's job,

**Technical Terms . . .**

**"Derivation"**

"The process of applying the rules of a grammar in succession in order to generate (or recognize) a "legal" sentence."

---

[92] Hausser, p. 176.
[93] Jurafsky points out some of the evidence used to suggest that humans may actually use context free languages, or at least constituency, in their mental processing of language. Cf., Jurafsky & Martin, p. 350-352.
[94] Dale, p. 4.
[95] Moisl, Comp. Ling. II, Lecture #4, p. 2.
[96] Samuelsson, Christer and Mats Wirén, "Parsing Techniques", in *Handbook of Natural Language Processing.* Dale, et. al. [eds.]. p. 63.

essentially is to search through all "possible" parse-trees that can be created from the grammar, and to find one that has *S* at its root and the exact words of the sentence at its leaves. If a sentence can be so "*derived,*" it is a grammatical sentence. (In some cases, there is more than one way to parse a sentence, that is, more than one parse tree exists for it; in that case the sentence is called "*ambiguous.*" We will discuss ambiguity in our section 4: Problems of NLP.)

Quite a number of different schemes and algorithms have been devised to accomplish this task, but all of the *rules*-based models can be divided into two essential classes: "*Top-Down*" and "*Bottom-Up.*"

Top-down parsers start at the root of the tree, and continually work their way down the tree, subdividing the non-terminal symbols into their constituent phrases until they reach the leaves. The goal is to find a "path" or a "derivation" from the root to the leaves. Conversely, a bottom-up parser starts with the sentence itself and "tries to build trees from the words up, again by applying rules from the grammar one at a time."[97] As Jurafsky explains, both methods have their advantages and disadvantages:

> The top-down strategy never wastes time exploring trees that cannot result in an *S*, since it begins by generating just those trees. . . . In the bottom-up strategy, by contrast, trees that have no hope of leading to an *S*, or fitting in with any of their neighbors, are generated with wild abandon. . . .
>
> The top-down approach has its own inefficiencies. While it does not waste time with trees that do not lead to an *S*, it does spend considerable effort on *S* trees that are not consistent with the input, . . . they generate trees before ever examining the input.[98]

No need to take sides on this issue, however, because the two most efficient algorithms employed in Syntax parsing adopt opposing approaches. The "*Earley*" algorithm is a top-down model, while the "*CYK*" is bottom-up. We will briefly discuss the Earley algorithm here, and leave the CYK as an exercise for the reader. ☺

The Earley parsing algorithm takes advantage of a creative concept known as "Dynamic Programming" (see the accompanying sidebar). This type of programming solves problems by "systematically filling in tables of solutions to sub-problems."[99] In this case, what the algorithm stores in the tables are *subtrees*. This avoids the problem that simpler parsers are tripped up by: namely, reparsing the same subtree over and over again. Such a redundant accumulation of parse trees can grow *exponentially* and render the problematic parsers effectively useless. The Earley, in contrast, implements a parallel top-down search that eliminates repetitious re-parsing or backtracking.[100]

---

[97] Jurafsky & Martin, p. 361.
[98] Ibid., p. 363.
[99] Jurafsky & Martin, p. 379

It might be noted that "the Earley algorithm uses the rules of the grammar, but not directly. Instead, the parsing algorithm disassembles the rules of the grammar successively into their basic elements, whereby the order is determined by the sequence of terminal symbols in the input string – and not by the logical derivation order of the grammatical rule system."[101]

## A Clever Idea . . .

**"Dynamic Programming"**

A special technique that is commonly used in many different phases of NLP is known as "Dynamic Programming." This is a special class of algorithms, first introduced by Bellman in 1957. As Jurafsky explains, "The intuition of a dynamic programming problem is that a large problem can be solved by properly combining the solutions to various subproblems."[102] The manner in which this is usually done, is by systematically filling in *tables* of *solutions to sub-problems*. When complete, the tables contain the solution to all the sub-problems needed to solve the problem as a whole.

For example, one procedure that makes use of this class of algorithms is the "minimum edit distance" algorithm for spelling error corrections. This is a technique that determines the fewest number of letters which must be changed to transform an incorrectly-spelled word to a correct word. To accomplish this, a matrix is created with the incorrect word spelled down the left column, and the correct "target" across the top; the cells are then filled in with the minimum number of changes that need to be made to transform the word at any given point:

|   | 1.1.1 | a | y | s |
|---|---|---|---|---|
| T | **1** | 2 | 3 | 4 |
| a | 2 | **1** | 2 | 3 |
| e | 3 | 2 | **2** | 3 |

---

[100] Ibid., p. 380.
[101] Hausser, p. 175.
[102] Jurafsky, p. 155. Cf. also, p.379.

> In this example, it takes only one letter substitution to change "Ta" to "Da" and 2 substitutions to change "Taes" to "Days". The value in each cell (the "minimum edit distance") can be computed as a simple function of the previously computed cells. Thus a difficult problem, can be broken down into smaller, easier problems "*dynamically.*"
> 
> As Jurafsky points out, dynamic programming is made use of by "the most commonly-used algorithms in speech and language processing, among them the *minimum edit distance* . . . the *Viterbi* algorithm and the *forward* algorithm which are used both in speech recognition and in machine translation, and the *CYK* and *Earley* algorithm used in parsing.[103]

### 3.2.4  Feature Structures & Unification

Despite the power of CFGs, as described above, they have contain a particular flaw if implemented simplistically. Consider this CFG, for example:

$$S \rightarrow NP\ VP$$
$$NP \rightarrow \text{The ball} \mid \text{The balls}$$
$$VP \rightarrow \text{roll} \mid \text{rolls}$$

This grammar correctly recognizes such sentences as "The ball rolls" and "The balls roll." Unfortunately, it also (incorrectly) recognizes "The ball roll " and "The balls rolls."  The problem has to do with a concept called "*agreement*" and it only gets more complex as one adds the complexities of *tense* (future, past, etc) and *person* ($1^{st}$, $2^{nd}$, $3^{rd}$).  In grammatically correct sentences, nouns and verbs (as well as determiners and auxiliaries) have to "*agree"* in person, tense and number.

How can we implement this constraint in the CFG model? A simplistic answer would be to add a rule for each and every case, but that would lead to an inordinate explosion of new rules such as:

$$S \rightarrow 3sgAux\ 3sgNP\ VP$$
$$S \rightarrow Non3sgAux\ Non3sgNP\ VP$$
$$3sgAux \rightarrow \text{does} \mid \text{has} \mid \text{can} \ldots$$
$$Non3sgAux \rightarrow \text{do} \mid \text{have} \mid \text{can} \ldots$$

---

[103] Ibid.

An alternative approach implements an entirely new device, called a "*feature structure."* Referring to these, Professor McTear points out

> Current grammatical formalisms in computational linguistics [i.e., formal theories regarding language processing], share a number of key characteristics, of which the main ingredient is a feature-based description of grammatical units. . . . These feature-based formalisms are similar to those used in knowledge representation and data type theory.[104]

These structures consist of so-called "*feature-value pairs,*" where "feature" refers to a category of syntax and "value" is the value that a particular word or phrase has taken within that category. Some simple examples might include:

[NUMBER = SG]

[PERSON = 3]

Every word or phrase in our input can now be associated with such a (set of) structures and the CFG parser can be instructed to combine two phrases only when their corresponding structures are in "agreement."

This description is, perhaps, overly-simplistic, and might leave the impression that feature sets are just an extension of the tagsets we had above. Indeed, with the example just given, such a comparison might be fair. But in actuality, feature sets are far more powerful and can be far more elaborate than what we have seen here. In particular, one of the ways feature sets may be examined, is by including one feature set as a *value* within another feature set; thus, information may be encoded in a hierarchical structure of embedded sets. These "sets of sets" can be described and analyzed using diagrams known as *Directed Acyclical Graphs*, or *DAGs*. These DAGs are charts which can be used to trace a path through a complex set of features, evaluating them and comparing them to the features of another set.

The process of comparing the feature structures of two constituent phrases, and combining them into one new phrase (and one new "unified" feature structure) is known as "*unification.*" Hence, "feature-based grammars are often subsumed under the term *unification grammars."* [105] Although this unification process is a little too complex to describe in totality here, it turns out that an algorithmic implementation of it is fairly easy. Jurafsky explains,

---

[104] McTear, p. 107.
[105] McTear, p. 107

"Roughly speaking, the algorithm loops through the features in one input and attempts to find a corresponding feature in the other. If all of the features match, then the unification is successful. If any single feature causes a mismatch then the unification fails".[106] The result is a CFG parser with significantly enhanced power, but without the huge proliferation of grammatical categories and rules.

Perhaps an example might help demonstrate some of the power of feature structures. Consider a particular NL implementation known as RASA -- a tool designed to enhance command and control capability in a military command post. At the center of Rasa is an NL engine that produces feature structures (attribute value pairs) as the method by which the *meaning* of a statement can be represented. Rasa happens to be a multi-modal processor which means that it not only deciphers spoken language, but also gestures, visual cues and physical tokens such as post-it notes on a map. David McGee and Philip Cohen, two of Rasa's designers, describe some of the relevant architecture:

> In Rasa, multimodal inputs are recognized, and then parsed, producing meaning descriptions in the form of typed feature structures. The integrator fuses these meanings together [through unification] by evaluating any available integration rules for the type of input received and those partial inputs waiting in the integration buffer.[107]

As you can see from the accompanying figure, the feature structures can grow to be quite complex. This should not be surprising however, because they represent complex entities. In the case of Rasa, a feature structure might represent a military unit. When an officer using Rasa speaks the words "Advanced Guard" the system generates a new feature structure and fills it in with all the characteristics about that particular unit that are known at the time. A feature structure such as this, which represents specific entities, is known as a *typed* feature structure.

---

[106] Jurafsky & Martin, p. 418.
[107] McGee, David & Philip Cohen, "Creating Tangible Interfaces by Augmenting Physical Objects with Multimodal Language," in *IUI 2001 Intern'l Conference: Proceedings.* New Mexico: ACM Press (2001): p. 117. The authors point out that "typed feature structure unification is ideal for multimodal integration because it can combine complementary or redundant input from different modes, yet it rules out contradictory inputs."

Once two or more such "units" are placed in the system (perhaps the result of *different* modes of input), the process of unification can be initiated. Unification begins on the right-hand side ("rhs") of the structure in the child or "daughter" nodes ("dtr"). Referring to the figure above, McGee explains this process:

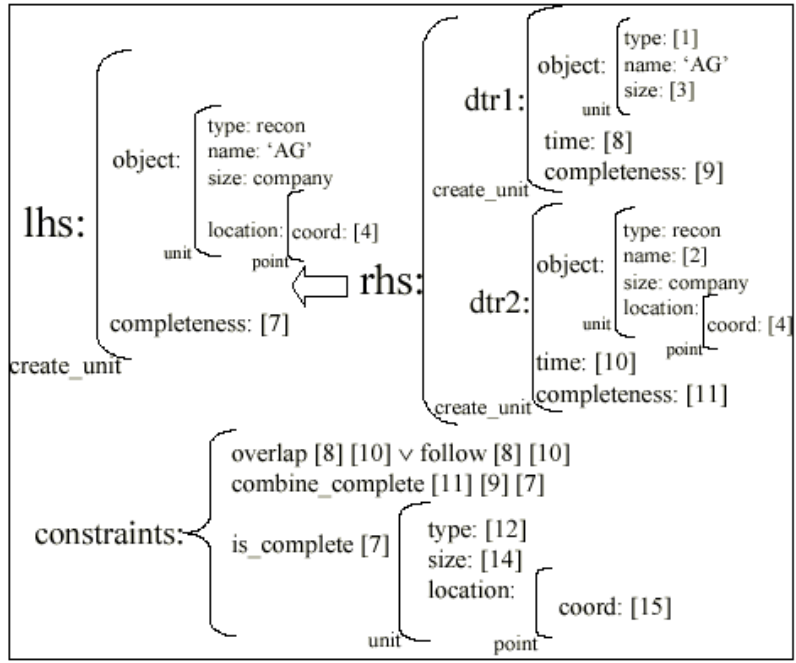

> Edges in the chart are processed by. . . grammar rules. In general these rules are productions LHS ← DTR1 DTR2 . . . DTRn; daughter features are fused, under the constraints given into thr left-hand side. The shared variable in the rules, denoted by numbers in square brackets, mustunify appropriately with the inputs from the various modalities. . . .
> 
> One of Rasa's multimodal grammar rules (see the figure) declares that partially specified units (`dtr1` and `dtr2`) can combine with other partially specified units, so long as they are compatible in type, size, location and name features, and they meet the constraints.[108]

Mcgee points out that the figure "demonstrates partial application of the rule and shows that after fusion the left-hand side is still missing a location feature for the unit specification."[109] In other words, Rasa doesn't know where on the map this unit is supposed to be located. And so, since this feature is missing, Rasa will produce a query to ask the officers where the unit is located.

From this brief example we can see that feature structures, though fairly simple in concept, can be put to powerful use by "intelligent" systems and offer some significant advantages.

McTear explains some other inherent values that feature structures offer:

> One major advantage of unification grammars is that they permit a declarative encoding of grammatical knowledge that is independent of any specific processing algorithm. A further advantage is that a similar formalism can be used for semantic representation, with the effect that the simultaneous use of syntactic and semantic constraints can improve the efficiency of the linguistic processing.[110]

---

[108] Ibid.
[109] Ibid.

This "syntactic and semantic" union was demonstrated in the Rasa implementation, where the structure was not only used for parsing, but for actually storing "*meaning."* We will have more to say about this in our next section.

We have now seen the process of Syntax Analysis and, of course, a great deal more could be said. It is clearly a valuable and powerful tool that has developed down a variety of different paths and has rooted itself as an indispensable intermediary between word recognition and meaning ascertainment. With it we identify "the underlying structure of a sequence of words. . . [providing] a structured object that is more amenable to further manipulation and subsequent interpretation."[111] And now, having derived the formal structure inherent in a given statement or utterance, we are finally prepared to input that tree of information into the final module of our NL processor: The Semantic Analyzer.

## 3.3  Semantic Analysis

The dividing line between Syntax and Semantics is often a blurry one. Both processes use many of the same tools and concepts in their respective goals of interpreting the text.  As one author explains,

> SYNTAX studies the structure of well-formed phrases (spelled out as sound sequences); SEMANTICS deals with the way syntactic structures are interpreted. However, how to exactly slice the pie between these two disciplines and how to map one into the other is the subject of controversy. In fact, understanding how syntax and semantics interact (i.e., their interface) constitutes one of the most interesting and central questions in linguistics.[112]

One real distinction that can be drawn between the two procedures, however, has to do with the underlying *goal* inherent in them: Syntax analysis is, essentially, the search for *structure*. Semantic analysis, on the other hand, is searching for *meaning.* And in particular, the goal is "to model how the meaning of an utterance is related to the meanings of the phrases, words and morphemes that constitute it.[113]

---

[110] McTear, p. 107-108.
[111] Dale, p. 5.
[112] Chierchia, Gennaro. "Syntax-Semantic Interface", in *MITECS,* http://cognet.mit.edu/MITECS/Entry/chierchia

Hausser elaborates,

> The ultimate goal, for humans as well as natural language-processing (NLP) systems, is to understand the utterance—which, depending on the circumstances, may mean incorporating the information provided by the utterance into one's own knowledge base or, more in general, performing some action in response to it.[114]

And, as Dale points out, "It is here that we begin to reach the bounds of what has managed to move from the research laboratory to practical application."[115] He explains some of the reasoning behind this:

> We know quite a lot about general techniques for . . . lexical analysis and syntactic analysis, but much less about semantics and discourse-level processing. . . . The known is the surface text, and anything deeper is a representational abstraction that is harder to pin down; so it is not so surprising that we have better-developed techniques at the more concrete end of the processing spectrum.[116]

That said, there has, nonetheless, been a great deal of successful research in this area, and significant breakthroughs continue to be achieved. This section will explore the goals and some of the significant features involved in such research.

### 3.3.1 Definitions: "The Meaning of Meaning"

Before we can examine this process in detail, we must first understand the semantic meaning of words like "understand," "semantic" and "meaning."

First, regarding the term semantics, one will find a diversity of definitions, depending on the application. As one scholar points out,

> It is not surprising that "semantics" can "mean" different things to different researchers within cognitive science. Notions relating to meaning

---

**Technical Terms . . .**

**"Meaning"**

"The meaning of an expression, as opposed to its form, is that feature of it which determines its contribution to what a speaker says in using it. Meaning conveyed by a speaker is the speaker's communicative intent in using an expression, even if that use departs from the expression's meaning. Accordingly, any discussion of meaning should distinguish speaker's meaning from linguistic meaning."[1]

[1] Kent Bach, MITECS

---

[113] Jurafsky & Martin, p. 499.
[114] Poesio, Massimo. "Semantic Analysis," in *Handbook of Natural Language Processing,* Robert Dale, et. al. (eds.), p. 93.
[115] Dale, p. 5.
[116] Ibid., p. 2.

have had long (and often contentious) histories within the disciplines that contribute to cognitive science, and there have been very diverse views concerning what questions are important, and for what purposes, and how they should be approached. And there are some deep foundational and methodological differences within and across disciplines that affect approaches to semantics.[117]

Hausser describes some of the ways this term is used in various fields of science:

In *linguistics,* semantics is a component of grammar which derives representations of meaning from syntactically analyzed natural surfaces. In *philosophy*, semantics assigns set-theoretic denotations to logical formulas in order to characterize truth and to serve as the basis for certain methods of proof. In *computer science*, semantics consists in executing commands of a programming language automatically as machine operations.[118]

If "semantics" is hard to pin down, "understanding" is even more so. Researchers are often at a loss to describe what it means in *humans*, let alone in computers. What do we mean by "understanding," and how do we know if we have achieved it? Can a computer ever really "understand" *anything?*

In his essay on this subject, researcher Massimo Poesio offers these comments:
Research in NLP has identified two aspects of 'understanding' as particularly important for NLP systems. Understanding an utterance means, first of all, knowing what an appropriate response to that utterance is. For example, when we hear the instruction in {3} we know what action is requested, and when we hear {4} we know that it is a request for a verbal response giving information about the time:

    {3}    Mix the flour with the water.

    {4}    What time is it?

An understanding of an utterance also involves being able to draw conclusions from what we hear or read, and being able to relate this new information to what we already know. For example, when we semantically interpret {5} we acquire information that allows us to draw some conclusions about the speaker; if we know that Torino is an Italian town, we may also make some guesses about the native language of the speaker, and his preference for a certain type of coffee. When we hear {6}, we may conclude that John bought a ticket and that he is no longer in the city he started his journey from, among other things.

    {5}    I was born in Torino.

---

[117] Partee, Barbara H. "Semantics," in MITECS. http://cognet.mit.edu/MITECS/Front/linguistics.html

{6}    John went to Birmingham by train. [119]

Clearly, requiring this kind of interpretation is a tall order, and one complicated by the fact that "there is little agreement among researchers on what exactly the ultimate interpretation of an utterance should be, and whether this should be domain-dependent."[120] The result of this disagreement is a divergence between research-level and production-level implementations.

> Thus, whereas researchers often try to design general-purpose parsers, practical NLP systems tend to use semantic representations targeted for the particular application the user has in mind and semantic interpreters that take advantage of the features of the domain.[121]

In the face of all this diversity of opinion, it is perhaps surprising that any progress has been made at all. But, in fact, a good deal of consensus does exist, and has enabled researchers to make remarkable headway in the pursuit of computational understanding. Indeed, the methodological differences have "partly impeded but also stimulated cooperative discussion and fruitful cross-fertilization of ideas, and there has been great substantive progress in semantics. . . in recent decades."[122] The following two sections will highlight some of the preeminent features of this research.

### 3.3.2    Meaning-Representation: "First-Order Predicate Calculus"

Before we can ever assign meaning to a statement, we must have a way to *represent* that meaning in a universal, standardized notation. Whatever notation we choose ought to be logical (i.e., demonstrably self-consistent), flexible (i.e., able to express a universe of different ideas), and computationally tractable (i.e., feasible to implement in static computer language).  In addition, as Jurafsky and Martin point out, a meaning-representation system must only generate statements that are:

---

[118] Hausser, p. 371.
[119] Poesio, p. 95.
[120] Ibid., p. 94.
[121] Ibid.
[122] Partee, "Semantics," in MITECS, http://cognet.mit.edu/MITECS/Front/linguistics.html.

> **Technical Terms . . .**
>
> **"First-order predicate calculus"**
>
> "A theory in symbolic logic that formalizes quantified statements such as "there exists an object with the property that..." or "for all objects, the following is true...". First-order logic is distinguished from higher-order logic in that it does not allow statements such as "for every property, the following is true..." or "there exists a set of objects such that...". Nevertheless, first-order logic is strong enough to formalize all of set theory and thereby virtually all of mathematics. It is the classical logical theory underlying mathematics. "[1]
>
> ---
> [1] Source: Wikipedia, http\\:www.wikipedia.com

- **Verifiable** – such that the state of affairs described by a representation can be compared to the state of affairs in some world as modeled in a knowledge base.

- **Unambiguous** – so that, regardless of any ambiguity in the raw input, the representations have a single, unambiguous interpretation.

- **Canonical** – so that a variety of differently-worded inputs that all mean the same thing, should have the same representation – the *"canonical* form."

- **Inference- enabling** – allowing a system to draw valid conclusions based on the meaning representation of inputs and its store of background knowledge.

- **Expressive** – adequately handling an extremely wide range of subject matter.[123]

One particular system that satisfies all of these constraints and has been widely adopted as the principle tool in this regard is known as "*First-Order Predicate Calculus,"* of *FOPC.* Practitioners explain that "the appeal of predicate calculus is its simplicity. In particular, it has a straightforward syntax that allows us to concentrate on the semantics of a knowledge-base rather than the details of the syntax."[124] It is also popular because "it has formalizations that are provably sound and complete (i.e., that allow us to deduce from a set of sentences in the language all and only the sentences that are consequences of those sentences according to the semantics specified in the foregoing.)"[125]

To understand FOPC, we must first grasp the concept and purpose of logical languages in general. Hausser offers this explanation:

> In logical semantics, a simple sentence like **Julia sleeps** is analyzed as a proposition which is either true or false. Which of these two values is denoted by the propositions depends on the state of the world relative to which the proposition is interpreted. The state of the world, called the model, is defined in terms of sets and set-theoretic operations.[126]

---

[123] cf. Jurafsky & Martin, pp.504-511.
[124] Birnbaum, Larry. "CS337: Intro to Semantic Information Processing" *Lecture Notes*.
[125] Poesio, p. 96.
[126] Hausser, p. 375.

There is a two-level relationship, Hausser explains, between (1) the textual or audible expressions of a language (called the *"surface"*) and (2) the semantic content (or *"model"*), as illustrated in the following diagram:

LEVEL I      logical language:    *sleep (Julia)*

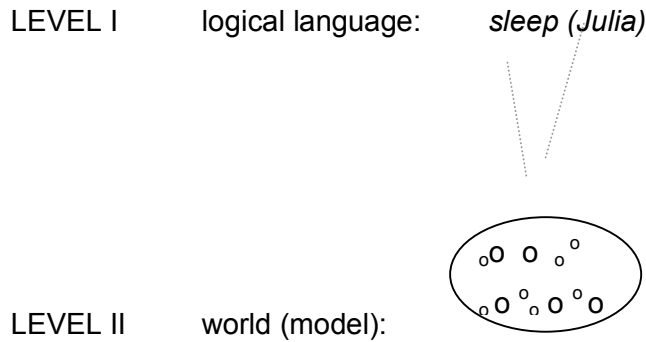

LEVEL II     world (model):

> By analyzing the surface **Julia sleeps** formally as *sleep(Julia)* the verb is characterized syntactically as a functor and the name as its argument. . . . In particular, the formal proposition *sleep(Julia)* is assigned the value true (or 1) relative to the model, if the individual denoted by the name is an element of the set denoted by the verb. Otherwise, the propositions denotes the value false (or 0).[127]

This basic ability to associate *predicates* ("*functors*" in Hausser) and their arguments is one of the central foundations of First Order Predicate Calculus. FOPC is essentially a "logic language" which is composed of things such as functions, predicates, constants, connectors and operators. Using these components, FOPC allows us to express certain properties of sets of objects. Thus, as Poesio explains,

> We can represent the information conveyed by natural language sentences stating that an object is a member of a certain set (say, John is a sailor) by means of a PREDICATE such as sailor, denoting a set of objects, and a TERM such as j, denoting John; the ATOMIC FORMULA sailor(j) expresses the statement. That formula is a syntactic way to express what we have called a semantic interpretation of the sentence.[128]

Here is a simple example, presented by Jurafsky, of a representation for *Maharani serves vegetarian food:*

    *Serves(Maharani, VegetarianFood)*

---

[127] Hawser, p. 376
[128] Poesio, p.96.

In this case *Serves* is what is called a "two-place predicate" holding between the objects denoted by the constants *Maharani* and *VegetarianFood*. Not surprisingly, the complexity of expressions can escalate as we combine statements using operators and connector. Consider, for example, this representation of *I have a car*:

$\exists x, y\ Having(x) \wedge Haver(Speaker, x) \wedge HadThing(y,x) \wedge Car(y)$

where ∃ (the "existential quantifier") can be read "*there exists,*" and the ∧ operator ("*and*") instructs us that this sentence will be true if all of its three component atomic formulas are true. One more example illustrates two more types of operators:

$\forall x VegetarianRestaurant(x) \rightarrow Serves(x, VegetarianFood)$

"For this sentence to be true, it must be the case that every substitution of a known object for *x* must result in a sentence that is true."[129] This latter example demonstrates the use of the "inference connective" (→) which is the basis for the *logic* within the theory. Without this connective, we can only make statements which are true or false. For some applications "representing this information is all we need; for example, a simple railway enquiry system may just need to know that the user wants to go from city A to city B on a given day."[130] More generally, however, as Poesio points out, we often want to be able to draw some inferences from the information.

> What makes first-order logic a logic is that it also includes a specification of the VALID conclusions that can be derived from this information (i.e., which sentences must be true given that some other sentences are true).[131]

Expressions such as the ones listed above may certainly look awkward and abstruse to the uninitiated observer upon first encounter. But in practice, this notational system is not difficult to become comfortable with. And, although it only has a small number of terms, operators, and connectors, it has proven to a vast power at expressing human expressions. Jurafsky includes pages of examples to demonstrate FOPC's ability to represent a variety of linguistic entities, such as Categories, Events, Time, Aspect, Beliefs, and Actions.[132]

Of course, FOPC is not without its problems. As Poesio explains:

---

[129] Jurafsky & Martin, p. 515, 518.
[130] Poesio, p. 96.
[131] Ibid.
[132] Jurafsky & Martin, pp. 522- 537.

One problem that is often mentioned is that inference with first-order logic is computationally expensive—indeed, in general, there is no guarantee that a given inference process is going to terminate. This suggests that it cannot be an appropriate characterization of the way humans do inferences, as humans can do at least some of them very quickly. Researchers have thus developed logics that are less powerful and, therefore, can lead to more efficient reasoning. Prolog is perhaps the best known example of a trade-off between efficiency and expressiveness. . . .

> For others, and especially for linguists, the problem with first-order logic is the opposite: it is not powerful enough. Formal semanticists, in particular, have argued for more powerful logics, either on the grounds that they provide more elegant tools for semantic composition, or that they can be used to model phenomena that cannot be formalized in first-order logic.[133]

Quite a number of alternative systems have been proposed and implemented in one system or another. We saw in the previous section that typed feature structures can be used with great power. In addition, you may recall from the beginning of this paper that Natural Language *itself* has been proposed as a computationally powerful and flexible meaning-representation system.[134] Nevertheless, FOPC continues to be the "best-known general-purpose theory of knowledge representation. . . providing a good illustration of what a theory of knowledge representation is meant to do."[135] And beyond that, it has become "a sort of lingua franca among researchers in knowledge representation."[136]

### A Clever Idea . . .

**Cycorp**

Doug Lenat, CEO of Austin-based *Cycorp,* is an artificial intelligence pioneer who is leading the human "memome" project, an effort to codify all the common sense in a person's head.[137] He and his team have labored 17 years ("600 person-years) to codify facts such as "Once people die, they stop buying things."[138]

Using a form of symbolic logic called "predicate calculus" to classify and show the properties of information in a standard way, they have assembled a knowledge base containing 3 million rules of thumb that the average person knows about the world, plus about 300,000 terms or concepts. Cyc now contains concepts like "If you are carrying a container that's open on one side, you should carry it with the open end up."

---

[133] Poesio, p. 97.
[134] Iwanska (p. 3) lists several: UNO, Montagovian syntax, KRISP, episodic logic, and SnePS.
[135] Poesio, p. 96.
[136] Poesio, p. 97.
[137] Anthes, Gary. "Computerizing Common Sense" *ComputerWorld* (Apr, 2002) p. 30.
[138] Ibid..

Lenat argues that such a "sub-stratum" of knowledge will be indispensable for successful NLP. He offers these sentences by way of example:

- Fred saw the plane flying over Zurich.
- Fred saw the mountains flying over Zurich.

Although the sentences are very similar, humans have little difficulty in recognizing that in the first sentence, "flying" probably refers to the plane, while in the second sentence, "flying" almost certainly refers to Fred. Traditional NL systems will have difficulty resolving this syntactic ambiguity, but because CYC knows that planes fly and mountains do not, it will be able to parse these sentences just as easily as a human. It's difficult to see how this could be done without relying on a large database of common sense.

Here's another example, involving pronoun disambiguation:

- The police arrested the demonstrators because *they* feared violence.
- The police arrested the demonstrators because *they* advocated violence.

Using commonsense knowledge to guide the interpretation process allows Cyc to deal with the these problems of ambiguity in natural language without having to rely solely on statistical techniques. This base is combined with a sophisticated NL system which has three components: the lexicon, the syntactic parser, and the semantic interpreter.[139] More information on these pieces can be found on the CYC-NL page.

Up until now, the only people adding knowledge to the database were a small group of logicians using an esoteric language called "CycL". Now, however, thanks to a recently released NL interface, millions of people can add their knowledge to Cyc. Because of the acceleration, Lenat expects to "be at 10 million assertions a year from now. . . . A typical person knows about 100 million things about the world. I see us crossing that point in five years. It's difficult to predict the course thereafter."[140]

---

[139] http://www.cyc.com/products2.html
[140] Anthes, p.30.

### 3.3.3 Semantic Composition: Assigning Interpretations to Utterances

We have now seen how we can break a sentence down into a syntactic structure – represented by a parse tree (Section 3.2); and we understand how we can beak "meaning" down into a semantic structure – represented in FOPC (section 3.3.1). The question remains, how do we map from one to the other? How are the interpretations and meaning representations actually assigned to or associated with the linguistic input?

One of the most widely-described approaches is a method known as "Syntax-Directed Semantic Analysis." The underlying assumption of this approach is referred to as "the principle of compositionality, that is, that *the meaning of a sentence is a function of the meaning of its parts.*[141] Of course, this does *not* mean that the meaning of a sentence is just the sum of the meanings of its words. By "*parts*" of a sentence we refer not just to the words, but to the syntactic relationships between the words. The semantic analysis, therefore, is grounded in the syntax, from which we get the name "Syntax *Directed* analysis." The manner in which we accomplish this is specified by what is known as the "rule-to-rule hypothesis." In general,

> Each syntactic rule has a corresponding semantic rule and the analysis of the constituent structure of the sentence will lead to the semantic analysis of the sentence as the meanings of the individual constituents identified by the syntactic analysis are combined.[142]

So by augmenting every syntax rule with an associated semantic rule, the interpretation of a statement can be generated as an immediate consequence of the syntax parsing process. We can illustrate this (where the notation "S.semantic" refers to the semantic information contained or stored in S), by the following grammar:

---

| CFG (Syntax Rules) | Associated Semantic Rules |
|---|---|
| S → NP VP | S.semantic := |
| NP → PN | NP.semantic := {PN.semantic} |
| VP → IV | VP.semantic := {IV.semantic} |
| PN → John | PN.semantic := {John} |
|     \| Sally | PN.semantic := {Sally} |
| IV → runs | IV.semantic := {runs(x)} |
|     \| eats | IV.semantic := {eats(x)} |

Note that the semantic interpretation rule associated with such productions as NP → N and VP → IP simply assigns the meaning of the child to the parent. The rule associated with S → NP VP "constructs a proposition by using the interpretation of the VP as a predicate and the interpretation of the NP as an argument."[143] In a similar manner, all the rules of a given grammar can be associated with semantic representations (usually in the form of FOPC or some other logical knowledge representation). In certain cases, extensions to FOPC are necessary, such as the "*lamda notation*" and the "*complex-term"* notation. These will not be discussed here.

Here, then, we can see that, through a fairly straight-forward process, we can actually associate "meaning" to any given rule in our syntactic grammar. Now, of course, we have demonstrated this only on the most simplistic rules; as the rules grow in complexity, the corresponding semantic associations can become entangled as well. What this means is that the job of the NLP *designer* is one fraught with mighty challenges. But once the system is actually designed, the implementational operation of it develops quite naturally.

---

In the remainder of the paper we will look at just a few of the ways that NLP is being applied, both locally in PNNL, as well as in other labs and commercial applications.

---

[143] Poesio, p. 101.

# A Clever Idea . . .

**Autonomy Corporation**

A prosperous, Cambridge-based software company is demonstrating that NLP techniques, properly applied, can be very lucrative. *Autonomy Corp,* headed by Cambridge graduate Michael Lynch, became profitable in Q1 2000, and it already has over $3.4 billion dollars in market capitalization.

Their mission is to "provide the software infrastructure that automates operations on semi-structured and unstructured information in any digital domain."[144] Translated, that means they enable computers to form an understanding of a piece of text, Web pages, e-mails, voice, documents and people automatically. Structured information (like a database on a server) has long been easy to operate on. *Un-*structured data, however, (like e-mails, recorded voice conversations, or archive news videos) are limited in that they can only be properly processed by people – manual editors who label or categorize the media by hand. This is labor-intensive, costly, inaccurate and by virtue of its manual aspect, does not scale.

Autonomy now offers a new alternative: automatic categorization, summarization, and hyperlinks (cross-referencing). For example, some editor might label a video "Bush speech; Dec 2000." But it would be more helpful to know the concepts that were discussed during the speech. Their software can tell you just that – automatically. Its strength lies in its use of high performance pattern matching algorithms and sophisticated probability theory – what Lynch calls "Bayesian math."

The heart of the approach, according to Lynch, is "understanding gist conversation. You don't have to use the words, and you don't have to get everything right, because you're listening to a conversation. If you were given a paragraph of a conversation--a transcript--and had I wiped out, let's say, every fourth word, you'd probably tell me what the conversation was about. Our applications are not about a few words in a sentence. They're about listening to a conversation."[145]

"You say, 'I'm interested in the causes of the Egypt airliner crash,' and [our software] starts playing the news. . . . Banks are starting to put this in now for their analyst briefings, so you can search all the analysts' briefings and hear the analysts talk No tags are put in; no closed captioning for it to work-it's based on speech."[146]

---

[144] http://www.autonomy.com/autonomy_v3/Content/Autonomy/Background
[145] Ibid., p. 44.
[146] Ibid., p. 43.

# 4. THE PRACTICAL APPLICATIONS:

In order to gain a better sense of the overall utility of NLP, it is important to look at some of the ways that it is being implemented and used. Accordingly, in this section we will enter two separate laboratories, and attempt to get a glimpse of the ways current researchers are grappling with and applying the notions of NLP.

## *4.1* Conversational Robots

Some of the more "high-profile" applications of NLP have not yet reached the stage of commercialization. In laboratories and Universities around the world one can find a wide gamut of exciting prototypes and promising designs, still just in developmental stages. One interesting theme found in many labs is the belief that domestic robots will someday be pervasive in our society—and "uninitiated users will need a way to instruct them to adapt to the [users'] needs."[147]

Dr. Roland Hausser, a researcher who has written a full-length text book on this endeavor, asserts:

> The development of speaking robots is not a matter of fiction, but a real scientific task. Remarkably, however, theories of language have so far avoided a functional modeling of the natural communication mechanism, concentrating instead on peripheral aspects such as methodology (behaviorism), innate ideas (nativism), and scientific truth (model theory).[148]

Accordingly, a number of groups have adopted the goal to "construct autonomous cognitive. . . robots which can communicate freely in natural language."[149] The motivation behind this goal is not simply the convenience that such machines would provide, but also to pursue theoretical advancements. As Dr. Hausser explains, "For theory development, the construction of talking robots is of interest because an electronically implemented model of communication may be tested both externally in terms of the verbal behavior observed, and internally via direct access to its cognitive states."[150]

---

[147] Lauria, Stanislo, et. al. "Training Robots Using Natural Language Instruction," *IEEE Intelligent Systems*. (Sept/Oct 2001): p. 38.

[148] Hausser, Roland. *Foundations of Computational Linguistics: Human-Computer Communication in Natural Language.* p. 1.

[149] Ibid., p. 1.

Pragmatically, however, these researchers agree that if robots will ever have a chance of being accepted by the popular society, they will need to be able to adapt to the specific needs of their users – and this mandates a strong utility in natural communication. The assumption is that most users will be naïve about computer language, and will thus be unable to personalize robots using standard programming methods.

Enabling free communication with robots, of course, is a non-trivial problem. On problem is that real-world instructions have such subtle levels of abstraction. Tim Oates, from the University of Massachusetts, gives an example:

> Consider the difference between asking a robot to *push* an object and asking a robot to *shove* an object. The meanings of push and shove are similar in that both involve *contact*… and the applications of *force*…. However, the meanings of these words differ in the magnitude and duration of the force that is applied. Depending on the [context], these two words have either the same semantic features, or similar but different features.[151]

The computer science departments at two Universities in particular (Edinburgh and Plymouth) have teamed together to devise a method which they call "Instruction-Based Learning" (IBL) which trains robots using NL instructions. Until now, "work on verbal communication with robots has mainly focused on issuing commands – that is, on activating pre-programmed procedures using a limited vocabulary."[152] In contrast, IBL seeks to employ "unconstrained language in a real-world robotic application," and one which emphasizes an actual "learning" process on the part of the robot.[153] The developers point out a number of potential advantages that such a system offers:

> Natural language can concisely express rules and command sequences. Also, because it uses symbols and syntactic rules, it is well suited to interact with robots that represent knowledge at the symbolic level. Such symbolic communication can help robots learn faster than those that learn at the sensory-motor association level.[154]

---

[150] Ibid., p. 3.
[151] Oates, Tim, et. al., "Toward Natural Language Interfaces for Robotic Agents: Grounding Linguistic Meaning in Sensors," *ACM 2000.* (Jan 2000): p 227.
[152] Stanislo, p. 39.
[153] Ibid., 38.
[154] Ibid.

In this particular application they start the process with a predefined set of knowledge. This "innate" knowledge consists of "primitive sensory-motor procedures, . . such as *turn_left* or *follow-the-road*."[155] These tasks are given names and the names constitute "symbols" which are then associated with programmed "actions." At that point, when a user explains a new procedure to the robot (involving several primitive actions) the IBL system actually writes a new piece of program code to execute it, names the code, and can then *reuse* that code in a growing accumulation of new symbols. Thus, the complexity of the robot grows (the robot "learns") every time it receives a new instruction.

Here is a sample dialogue that was used to instruct the robot to go from the museum to the library:

> **User:** Go to the library.
> **Robot:** How do I go to the library?
> **User:** Go to the Post Office, go straight ahead, the library is on your left.[156]

In this example, the robot solicited an explanation to clarify the original instruction. The user's reply makes use of a previously learned symbol: the route to the Post Office.

Of course, interaction with this level of sophistication requires a solid grounding in NL processing. The IBL system is divided into four subtasks:

- speech recognition
- linguistic analysis
- ambiguity resolution, and
- dialogue updating.

In order to realize these subtasks, the designers have implemented a "dialogue manager." As illustrated in the accompanying figure,

> the dialogue manager converts speech input into a semantic representation and converts robot manager requests into user dialogue. The system runs its components as different processes communicating with each other through a blackboard architecture. The robot manager listens to new messages from the dialogue manager while processing previous ones using a multithread approach. It does this by launching a message-evaluation thread, **execution process**, through its communication interface. It then resumes listening to the dialogue manager. The **execution process** thread's aim is to understand the dialogue manager's message and act accordingly.[157]

---

[155] Ibid.
[156] Ibid., p. 40.

Each of these systems is described in fascinating detail in the IEEE publication cited below, and more information is obtainable from the group's Digital Library located at http://computer.org/publications/dlib.

This is just one example of the growing number of groups that believe "truly sentient robots need learning abilities that constrain abstract reasoning in relation to dynamically changing external events and the results of their own actions."[158] As Dr. Hausser explains, "For practical purposes, unrestricted communication with . . . robots in natural languages will make the interaction with these machines maximally user friendly and permit new, powerful ways of information processing."[159]

Unanswered questions remain, however, such as "Can this approach be generalized to other instruction contexts?" and "Can this approach create effective and socially acceptable robots?" One of the proposed experiments to help explore these questions is embodied in their next project: an autonomous wheelchair[160]. . . . Stay tuned.

## 4.2  Interview with David McGee,
## Pacific Northwest National Laboratory

David McGee is a Senior Research Scientist at Pacific Northwest National Laboratories in Richland, WA.  He is one of the Lab's leading researchers in a field of Computer-Human Interaction known as "Multi-Modal Interfaces," a subset of the "Rich Interaction Environments" Group.  Dr. McGee has been involved in two separate research and development projects which incorporate NL processing and generation: "Quickset" & "RASA."  His particular interest and experience lie in the specific field of "Command and Control" NLP. This is related to, but distinct from the "phone-based" applications (banking machines, etc.). "With phone systems, the input is entirely spoken, and the feedback is singular [in modality] as well." McGee's systems, in contrast, demand facilities with *multi-*modal input and *multi-*media output. They are not only listening to verbal input, but looking for visual cues (like gestures) as well. "The focus of my Ph.D. research, therefore, has tended to be quite a bit *broader* than most people's." This breadth of exposure has enlightened McGee to certain problems that exist commonly in current R&D.

---

[157] Ibid., pp. 41-42.
[158] Ibid., p. 44.
[159] Hausser, p. 3.
[160] Stanislo., p.44.

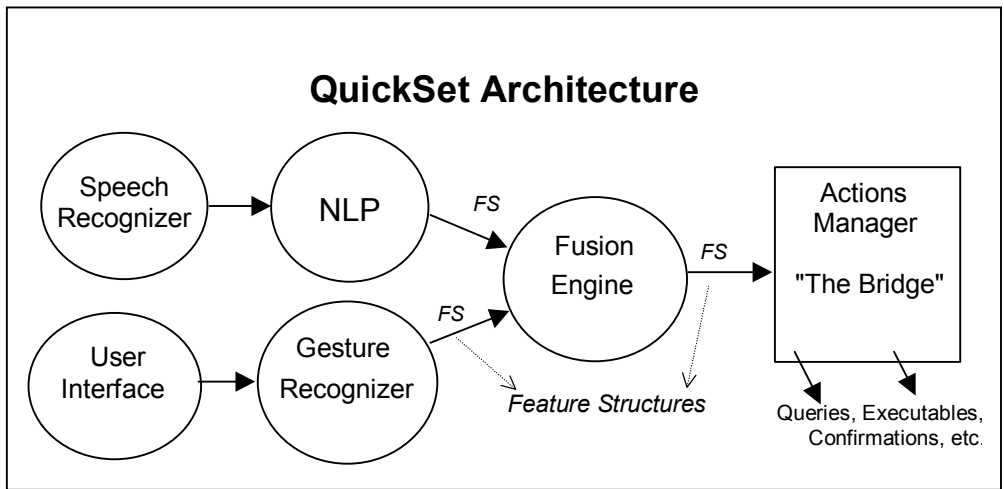

**QuickSet Architecture**

One of the concerns he has regarding the traditional approaches to NLP has to do with architectural "scope." He explains, "Most developers think that once the NL processor has produced its output, then that's it. That might be true with argumentation or summarization, where you evaluate the text and then you're done. But when you're thinking in terms of broad architecture, the issues become more complex." By way of example, he sketched a diagram of the QuickSet architecture.

Notice that the output from the NLP module are (the standard) feature structures. And for many NLP developers that's the end of the story. But a significant problem arises when this module is intertwined with others. In this example the NLP provides one source of input, the User Interface and the Gesture Recognizer provide another. Both of these outputs are fed into the Fusion engine, which, similarly, produces feature structures as output. Finally an Actions manager (called "the Bridge") determines the appropriate measures to take. One particular action that may be chosen is a "confirmation" with the user that the interpretation is correct. This is where a significant liability in the traditional model is exposed: the feature structures output from the NL processor (as well as from the Fusion engine) are (as in most typical implementations) in the form of *singletons.* If the Bridge attempts to confirm the accuracy of this output and the User does not confirm, the process is effectively dead. No other options are provided. "The NL processor assumed it was right – 'this *is* the interpretation.' But the user disagrees, so now we're stuck. Instead what we needed was some sort of *'N*-best' list of potential executables so the confirmation process has something to work with."[161]

    This is just one of the drawbacks of the standard architectural models, and it is the result, one could presume, of the narrow field of development that most researchers are involved with. It is not until a broader, longer-term perspective is implemented that such short-comings are revealed.

After discussing his own experience with NLP, I asked him to assess the current and future contributions that he believes NLP can and will have at PNNL. As McGee explains, PNNL already considers NL research to be a central key in their IS&E projects. They are pursuing and developing many different "methods to enable users to interact naturally with computers that are not necessarily visible." McGee says, "We want to understand what intuitions are involved when people interact with technology, and how we might guide those intuitions. This involves understanding people's intentions as they interact in natural ways – both verbally and with gestures." In addition, the Lab has a large investment in textual processing. "But," McGee continues, "we are recognizing that the statistical approaches are inherently flawed… as are the rules-based approaches. We have to *combine* the two."

---

[161] All the quotes in this section were taken from Dr. David McGee, in an interview at PNNL on     April 30, 2002.

# 5. CONCLUSION:

We have finally attained the apex of our goal: we have seen the process of translating an utterance into computational code. We have traced the path of an input statement through lexical analysis into string of words, then through syntax analysis into a parsed tree of feature structures, and finally through semantic analysis into a formally standardized knowledge representation. Of course, the many details we left out or skimmed over could fill many volumes (and they do!). But our attempt to grasp a high-level view of the NL process is complete. NL research in general, on the other hand, is far from complete.

A great deal more remains to be discovered and standardized. As Moens has said, "Doing a full semantic analysis of an input text is still a very complex undertaking. Despite several decades of research and development work, no practically usable, domain-independent parser of unrestricted text has been developed so far."[162]

And as Huang and Furui point out,

> To accomplish the ultimate goal of a machine that can communicate with people. . . a number of research issues are awaiting further study. Such a communicating machine needs to be able to deliver a satisfactory performance under a broad range of conditions and have an efficient way of representing, storing, and retrieving "knowledge" required in a natural conversation.[163]

"Fulfilling these requirements," as Hausser asserts, "will take hard, systematic, goal oriented work, but it will be worth the effort."[164] And, say Huang and Furui, "with the current enthusiasm in research advances, we [can be] optimistic that the Holy Grail of natural human-machine communication will soon be within out technological reach."[165]

How will such a technological revolution change society? We can only imagine. Dougherty exposed his optimism a few years ago as he speculated:

> [If we can] represent English, French, and German on a computer. . . then one will not have to learn to program in computer languages (C, Fortran, Cobol). Instead one could simply type in English sentences. If human languages were represented on a computer, then one could translate from one language to another automatically. On a more profound note, any problem that can be posed in any formal notation or in any formal language can be posed in English. If there was a computer that could correctly interpret English sentences, then it would correctly interpret any problem and give answers in unambiguous grammatical English sentences.[166]

---

[162] Moens, p. 4/4.
[163] Hwang & Furui, p. 1164.
[164] Hausser, p. 3.
[165] Hwang & Furui, p. 1164.

Perhaps dreams such as this will prove to have been too lofty. Or perhaps not. Who can tell what this technology will bring? One thing is clear, though. Within the next five to ten years, we will begin to see products emerging from laboratories around the world that will radically reshape how we as humans interact with information: we will finally be able to *talk* to our computers.

Now let's just hope they don't give us any lip!

---

[166] Dougherty, *Natural Language Computng* New York: University Press (1994): p. xxiii.

# ANNOTATED BIBLIOGRAPHY

(note: NL= Natural Language; NLP = NL Processing; NLG = NL Generation)

## 1. Textbooks:

Huang, Xuedong, et. al. *Spoken Language Processing: A Guide to Theory, Algorithm, and System Development.* New Jersey: Prentice Hall, 2002.
    This book draws on the latest advances and techniques from multiple fields: computer science, electrical engineering, acoustics, linguistics, mathematics, psychology and more. The authors (scientists at Microsoft Research) present detailed case studies base on stat-of-the-art systems, inluding Microsoft's Whisper speech recognizer, Whistler text-to-speech system, Dr. Who dialogue system, and the Mpad handheld device.

Iwañska, Łucja M. and Stuart C. Shapiro. Natural Language Processing and Knowledge Representation: Language for Knowledge and Knowledge for Language. Cambridge: The MIT Press: 2000.
    The authors of this book believe that NL has a significant and heretofore unexplored role in human information and knowledge processing. The contributors to the book are interested particularly in knowledge representation and reasoning systems — foundational elements in Artificial Intelligence (AI). They argue that only when computational models are driven by NL will they be able to develop into truly intelligent computer systems that simulate human knowledge processing. This thesis is admitted to be controversial, and the vast majority of knowledge representation and reasoning systems do not adequately reflect important characteristics of NL.
    The contrarian experts in the field of AI argue that NL is too inherently ambiguous and "non-algorithmic" to be of any use in knowledge processing; language is considered to be merely the "froth on the surface of thought" (p.xv), and NLP will forever be limited to "interfaces" to the real systems. The authors, on the other hand, believe that a viable case can be made for considering NL to be the best model of the structure of the human mind itself. In other words, "NL is essentially the language of human thought." Clearly, the conclusion to this debate will have significant ramifications for the future of NLP.
    Another helpful contribution of the book is its overview of several of the most cutting edge NL systems. In fact, the contributers to the book have each developed very mature and high level NLP systems, and this allows a helpful comparison to the various approaches in current research. (There is also a helpful appendix illustrating some of the remaining challenges in NLP research).

Jacquemin, Christian. Spottting and Discovering Terms through Natural Language Processing. Cambridge: The MIT Press, 2001.
    Jacquemin presents an extended study in one of the specialized areas of NLP: information retrieval from electronic documents. He is aiming for the goal of qualitative and accurate information retrieval that goes far beyond the simplistic and often erroneous word-searching utilities that are common today.
    In particular, Jacquemin describes a NL processing system known as FASTR. This system is a unique combination of several natural language processing techniques. It serves both as an efficient parser, capable of processing large amounts

of terminological and textual data, and as an accurate term spotter, capable of distinguishing subtle linguistic differences between correct and spurious term occurrences. His presentation of FASTR seeks to demonstrate that no in-depth understanding of target documents is necessary for a correct identification of terms and variants. Instead, this so-called "shallow parser" performs "text-skimming through *term identification*" in automated information retrieval.

Jurafsky, Daniel, and James H. Martin. Speech and Language Processing: An Introduction to Natural Language Processing, Computational Linguistics, and Speech Recognition. New Jersey: Prentice Hall, 2000.

    This is certainly the most important text in my study. It is considered indispensable reading, and is the primary text in many if not most university courses on NLP (and is considered by some to be the most influential NLP book of the last decade). The authors offer a unified vision of speech and language processing, presenting state-of-the-art algorithms and techniques for both speech and text-based processing of natural language. This comprehensive work covers both statistical and symbolic approaches to language processing; it shows how they can be applied to important tasks such as speech recognition, spelling and grammar correction, informational extraction, search engines, machine translation, and the creation of spoken-language dialog agents. The authors include an emphasis on scientific evaluation as well as practical applications. They cover all the new statistical approaches, while still covering the earlier more structured rule-based methods.

    The four sections of his book take the reader from the processing of Words, to Syntax, to Semantics, concluding with a segment on Pragmatics. Although topics such as speech recognition and spelling correction are outside the scope of my research, the bulk of this book is a clearly organized, highly readable overview of the entire science of NL.

Manning, Christopher D., and Hinrich Schütze. *Foundations of Statistical Natural Language Processing.* Cambridge: The MIT Press, 1999.

    This book is one of the two most commonly required texts in university NL courses. It represents the newest of the two major approaches to NLP, namely, statistical processing. Some argue that this has become the dominant approach in recent years, and this is the first comprehensive introductory text to appear on the subject.

    The book contains all the theory and algorithms needed for building NLP tools. It provides broad but rigorous coverage of mathematical and linguistic foundations, as well as detailed discussion of statistical methods. It covers collocation finding, word sense disambiguation, probabilistic parsing, information retrieval, and other applications. The book's layout is clear: after a section on preliminaries it addresses the three main issues of Words, Grammars, and Applications or Techniques.

    The book was intended to be used in a semester-long graduate-level course in Statistical NLP, and as such covers much more information than can feasibly be digested in my current study. Nevertheless, it clearly offers a wealth of the most current information in this burgeoning field.

Niyogi, Partha. The Informational Complexity of Learning: Perspectives on Neural Networks and Generative Grammar. Boston: Kluwer Academic Publishers, 1998.

Niyogi, an expert in the enormously complex process of learning, asks the question: how do humans (or, by extension, computers) acquire knowledge? The presumption is that learning is the centerpiece of human intelligence, and any attempt to replicate intelligence in computers must first explain this remarkable ability.

Although this book delves deeply into concepts which are tangential to my current topic (e.g., psychological and statistical learning theories, etc), it contains two chapters (chs 4 & 5) that deal directly with the issue of NL: the process of learning natural language grammars. The central algorithm explored (the "Triggering Learning Algorithm" or TLA) is suggested as the best model for understanding this process.

Aho, Alfred V., et. al. *Compilers: Principles, Techniques, and Tools*. Reading, MA: Addison-Wesley, 1988.

Albert, Dietrich & Josef LukasMahwah, *Knowledge Spaces: Theories, Empirical Research, and Applications.* N.J. : Lawrence Erlbaum Associates, Inc., 1999.

Dale, Robert, et. al. (eds). *Handbook of Natural Language Processing.* Marcel Dekker, 2000.

This extensive 900-page tome was tragically not stocked at any of the libraries avialable to me. (Perhaps not surprising with its $200 price tag). It appears, however, to be a very valuable resource, and one which must clearly be included in any further research on this topic.

Dougherty, Ray. *Natural Language Computing : An English Generative Grammar in Prolog.* New York: Lawrence Erlbaum Assoc, 1994.

Hausser, Roland. *Foundations of Computational Linguistics: Human-Computer Communication in Natural Language.* New York: Springer, 2001.

## 2. Conference Proceedings / Published Lecture Notes:

Harras, Gisela. "Concepts in Linguistics – Concepts in Natural Language." Conceptual Structures: Logical, Linguistic, and Computational Issues: 8[th] Int'l Conference on Conceptual Structures, Proceedings. Ed. Bernhard Ganter and Guy W. Mineau. Berlin: Springer, 2000. Pp. 13-26.

Harras grapples with one of the difficulties in interpretation: mapping a word or lexeme (a "sign", in his terminology) with the external referent or physical/conceptual reality. Essentially this is a discussion about the ambiguities of communication. An example given compares the sentences: "John left the institute an hour ago" and "John left the institute a year ago." That human cognitive systems have no difficulty with interpreting these statements is a sign that "the lexicon of a language, conceptual knowledge, and communicative acting are all inseperately interrelated." This, of course, is in contrast to artificial systems.

Although this portion of the discussion looks interesting, the remainder of the article delves deeply into cognitive theory and, in particular, the arcane debate between "one-level-" and "two-level-semantics." These are issues that some NL theoriticians will, no doubt, have to wrestle with at some point, but it seems somewhat far afield from my current line of research.

**3. Journals and Articles:**

language understanding, information retrieval, language generation and speech synthesis. An interesting, but all-too-brief description of the application is included.